\documentclass{article}
\usepackage{amsmath,epsfig,bm, amssymb,subfigure,natbib}

\newcommand{\tr}[1]{\mathrm{tr}(#1)}

\newcommand{\inner}[2]{\langle #1,#2\rangle}

\newcommand{\mathbbR}{\mathbb{R}}

\newcommand{\boldone}{{\boldsymbol{1}}}

\newcommand{\boldH}{{\boldsymbol{H}}}
\newcommand{\boldI}{{\boldsymbol{I}}}

\newcommand{\boldK}{{\boldsymbol{K}}}
\newcommand{\boldL}{{\boldsymbol{L}}}

\newcommand{\boldh}{{\boldsymbol{h}}}

\newcommand{\boldalpha}{{\boldsymbol{\alpha}}}
\newcommand{\boldbeta}{{\boldsymbol{\beta}}}

\newcommand{\boldtheta}{{\boldsymbol{\theta}}}

\newcommand{\boldGamma}{{\boldsymbol{\Gamma}}}

\newcommand{\calE}{{\mathcal{E}}}
\newcommand{\calF}{{\mathcal{F}}}
\newcommand{\calG}{{\mathcal{G}}}

\newcommand{\calX}{{\mathcal{X}}}

\newcommand{\calZ}{{\mathcal{Z}}}

\newcommand{\domainx}{\calX}

\newcommand{\densitysymbol}{p}

\newcommand{\density}{\densitysymbol}

\newcommand{\nsample}{n}

\usepackage{graphicx}

\advance\oddsidemargin-1in
\date{\today}
\title{Least-Squares Independence Regression for Non-Linear Causal Inference under Non-Gaussian Noise} 

\author{Makoto Yamada$^1$, Masahi Sugiyama$^{1,2}$, and Jun Sese$^3$\\
$^1$Tokyo Institute of Technology, Japan\\
$^2$PRESTO, Japan Science and Technology Agency, Tokyo, Japan\\
$^3$Department of Information Science, Ochanomizu University, Japan\\
\texttt{\{yamada@sg. \hspace{-3mm}sugi@\}cs.titech.ac.jp}, \texttt{sesejun@is.ocha.ac.jp}}

\begin{document}
\maketitle

\begin{abstract}
The discovery of non-linear causal relationship under additive non-Gaussian noise models
has attracted considerable attention recently because of their high flexibility.
In this paper, we propose a novel causal inference
algorithm called {\em least-squares independence regression} (LSIR).
LSIR learns the additive noise model through the minimization of 
an estimator of the \emph{squared-loss mutual information}
between inputs and residuals.
A notable advantage of LSIR over existing approaches is that 
tuning parameters such as the kernel width and the regularization parameter
can be naturally optimized by cross-validation,
allowing us to avoid overfitting in a data-dependent fashion.
Through experiments with real-world datasets, 
we show that LSIR compares favorably with a state-of-the-art 
causal inference method.
\end{abstract}

\section{Introduction}
Learning \emph{causality} from data is one of the important challenges
in the artificial intelligence, statistics, and machine learning communities
\citep{book:Pearl:2000}.
A traditional method of learning causal relationship from observational data
is based on the linear-dependence Gaussian-noise model \citep{UAI:Geiger+Hackerman:1994}.
However, the linear-Gaussian assumption is too restrictive
and may not be fulfilled in practice.
Recently, non-Gaussianity and non-linearity have been shown to
be beneficial in causal inference,
allowing one to break symmetry between observed variables
\citep{JMLR:Shimizu+etal:2006,NIPS:Hoyer+etal:2009}.
Since then, much attention has been paid to
the discovery of non-linear causal relationship through non-Gaussian noise models
\citep{ICML:Mooij+etal:2009}.

In the framework of non-linear non-Gaussian causal inference,
the relation between a cause $X$ and an effect $Y$ is assumed to be described by
$Y = f(X) + E$,
where $f$ is a non-linear function and $E$ is non-Gaussian additive noise
which is independent of the cause $X$.
Given two random variables $X$ and $X'$,
the causal direction between $X$ and $X'$ is decided
based on a hypothesis test of whether the model $X' = f(X) + E$
or the alternative model $X = f'(X') + E'$ fits the data well---here,
the goodness of fit is measured by independence between inputs
and residuals (i.e., estimated noise).
\citet{NIPS:Hoyer+etal:2009}
proposed to learn
the functions $f$ and $f'$ by the \emph{Gaussian process} (GP)
\citep{book:Bishop:2006}, and 
evaluate the independence between the inputs and
the residuals by 
the \emph{Hilbert-Schmidt independence criterion} (HSIC)
\citep{ALT:Gretton+etal:2005}.

However, since 
standard regression methods such as GP
are designed to handle Gaussian noise,
they may not be suited for discovering
causality in the non-Gaussian additive noise formulation.
To cope with this problem, a novel regression method called {\em HSIC regression} (HSICR)
has been introduced recently \citep{ICML:Mooij+etal:2009}.
HSICR learns a function so that the dependence between inputs and residuals is
directly minimized based on HSIC.
Since HSICR does not impose any parametric assumption
on the distribution of additive noise, it is suited
for non-linear non-Gaussian causal inference.
Indeed, HSICR was shown to outperform the GP-based method
in experiments \citep{ICML:Mooij+etal:2009}.

However, HSICR still has limitations for its practical use.
The first weakness of HSICR is that the kernel width of HSIC needs to be determined manually.
Since the choice of the kernel width heavily affects the sensitivity
of the independence measure \citep{TAS:Fukumizu+etal:2009}, 
lack of systematic model selection strategies is critical in causal inference.
Setting the kernel width to the median distance between sample points
is a popular heuristic in kernel methods \citep{book:Schoelkopf+Smola:2002},
but this does not always perform well in practice.
Another limitation of HSICR is that the kernel width of the regression model
is fixed to the same value as HSIC.
This crucially limits the flexibility of function approximation in HSICR.

To overcome the above weaknesses, we propose an alternative regression method
called {\em least-squares independence regression} (LSIR).
As HSICR, LSIR also learns a function so that the dependence
between inputs and residuals is directly minimized.
However, a difference is that,
instead of HSIC, LSIR adopts an independence criterion called
{\em least-squares mutual information} (LSMI) \citep{BMCBio:Suzuki+etal:2009a},
which is a consistent estimator of the \emph{squared-loss mutual information} (SMI)
with the optimal convergence rate.
An advantage of LSIR over HSICR is that
tuning parameters such as the kernel width and the regularization parameter
can be naturally optimized through cross-validation (CV) with respect to 
the LSMI criterion.

Furthermore, we propose to determine the kernel width of the regression model
based on CV with respect to SMI itself.
Thus, the kernel width of the regression model
is determined independent of that in the independence measure.
This allows LSIR to have higher flexibility in non-linear causal inference than HSICR.
Through experiments with real-world datasets,
we demonstrate the superiority of LSIR.

\section{Dependence Minimizing Regression by LSIR}
In this section, we formulate the problem of dependence minimizing regression and propose a novel regression method, {\em least-squares independence regression} (LSIR).
\subsection{Problem Formulation}
\label{sec:formulation}
Suppose random variables $X\in\mathbbR$ and $Y\in\mathbbR$
are connected by the following additive noise model \citep{NIPS:Hoyer+etal:2009}:
\[
Y = f(X) + E,
\]
where  $f: \mathbbR \rightarrow \mathbbR$ is some non-linear function
and $E \in \mathbbR$ is a zero-mean random variable independent of $X$. 
The goal of dependence minimizing regression
is, from i.i.d.~paired samples $\{(x_i, y_i)\}_{i = 1}^n$,
to obtain a function $\widehat{f}$ such that input $X$
and estimated additive noise $\widehat{E} = Y - \widehat{f}(X)$ are independent. 

Let us employ a linear model for dependence minimizing regression:
\begin{align}
  f_{\boldbeta}(x) = \sum_{l = 1}^m \beta_l \psi_l(x)
             = \boldbeta^\top{\bm \psi}(x),
             \label{regression-model}
\end{align}
where $m$ is the number of basis functions,
$\boldbeta=(\beta_1,\ldots,\beta_m)^\top$ are regression parameters, $^\top$ denotes the transpose,
and ${\bm \psi}(x)=(\psi_1(x),\ldots,\psi_m(x))^\top$ are basis functions.
We use the Gaussian basis function in our experiments:
\[
\psi_l(x)= \exp \left(-\frac{ (x - c_l )^2}{2\tau^2}\right),
\]
where $c_l$ is the Gaussian center chosen randomly from 
$\{x_i\}_{i = 1}^n$ without overlap
and $\tau$ is the kernel width.

In dependence minimization regression, we learn the regression parameter $\boldbeta$ as
\[
\min_{\boldbeta} \left[I(X, \widehat{E}) + \frac{\gamma}{2} \boldbeta^\top \boldbeta  \right],
\]
where
 $I(X,\widehat{E})$ is some measure of independence between $X$ and $\widehat{E}$, 
 and $\gamma\ge0$ is the regularization parameter for avoiding overfitting. 

In this paper, we use the \emph{squared-loss mutual information} (SMI)
\citep{BMCBio:Suzuki+etal:2009a} as our independence measure:
\begin{align*}
\textnormal{SMI}(X, \widehat{E}) &=
\frac{1}{2}\iint \left( \frac{p(x, \widehat{e})}{p(x)p(\widehat{e})} - 1\right)^2
 p(x)p(\widehat{e})\textnormal{d}x \textnormal{d}\widehat{e}.
\end{align*}
$\textnormal{SMI}(X, \widehat{E})$ is the \emph{Pearson divergence}
\citep{PhMag:Pearson:1900} from $p(x, \widehat{e})$
to $p(x)p(\widehat{e})$,
and it vanishes if and only if $p(x, \widehat{e})$
agrees with $p(x)p(\widehat{e})$,
i.e., $X$ and $\widehat{E}$ are independent.
Note that ordinary \emph{mutual information} (MI) \citep{book:Cover+Thomas:2006},
\begin{align}
\textnormal{MI}(X, \widehat{E}) &=
\iint p(x, \widehat{e})\log\frac{p(x, \widehat{e})}{p(x)p(\widehat{e})}
\textnormal{d}x \textnormal{d}\widehat{e},
\end{align}
corresponds to the \emph{Kullback-Leibler divergence} \citep{Annals-Math-Stat:Kullback+Leibler:1951}
from $p(x, \widehat{e}) $ and $p(x)p(\widehat{e})$,
and it can also be used as an independence measure.
Nevertheless, we adhere to using SMI since it allows us to obtain an analytic-form estimator,
as explained below.

\subsection{Estimation of Squared-Loss Mutual Information}
SMI cannot be directly computed since it contains unknown densities
$p(x, \widehat{e})$, $p(x)$, and $p(\widehat{e})$.
Here, we briefly review an SMI estimator called \emph{least-squares mutual information} (LSMI)
\citep{BMCBio:Suzuki+etal:2009a}.

Since density estimation is known to be a hard problem
\citep{book:Vapnik:1998},
avoiding density estimation is critical for obtaining better SMI approximators
\citep{PhyE:Kraskov+etal:2004}.
A key idea of LSMI is to directly estimate the \emph{density ratio}:
\[
r(x, \widehat{e}) = \frac{p(x, \widehat{e})}{p(x)p(\widehat{e})},
\]
without going through density estimation of $p(x, \widehat{e})$, $p(x)$, and $p(\widehat{e})$.

In LSMI, the density ratio function $r(x, \widehat{e})$ is
directly modeled by the following linear model:
\begin{eqnarray}
r_{\boldalpha}(x, \widehat{e})= \sum_{l = 1}^b \alpha_l \varphi_l(x,\widehat{e})
 = \boldalpha^\top {\bm \varphi}(x, \widehat{e}),
\end{eqnarray}
where $b$ is the number of basis functions,
$\boldalpha=(\alpha_1,\ldots,\alpha_b)^\top$ are parameters,
and
${\bm \varphi}(x,\widehat{e})=(\varphi_1(x,\widehat{e}),\ldots,\varphi_b(x,\widehat{e}))^\top$
are basis functions.
We use the Gaussian basis function:
\[
\varphi_l(x,\widehat{e})= \exp \left(-\frac{ (x - u_l)^2
+ (\widehat{e} - \widehat{v}_l)^2}{2\sigma^2}\right),
\]
where $(u_l,\widehat{v}_l)$ is the Gaussian center chosen randomly from 
$\{(x_i,\widehat{e}_i)\}_{i = 1}^n$ without replacement,
and $\sigma$ is the kernel width.

The parameter $\boldalpha$ in the model $r_{\boldalpha}(x,\widehat{e})$ is learned
so that the following squared error $J_0(\boldalpha)$ is minimized:
\begin{align*}
J_0(\boldalpha) &= \frac{1}{2}\iint(r_{\boldalpha}(x,\widehat{e}) - r(x,\widehat{e}))^2
p(x)p(\widehat{e})\textnormal{d}x\textnormal{d}\widehat{e} \nonumber \\
&= \frac{1}{2}\iint r_{\boldalpha}(x,\widehat{e})p(x)p(\widehat{e})
\textnormal{d}x\textnormal{d}\widehat{e} - \iint r_{\boldalpha}(x,\widehat{e})p(x,\widehat{e})
\textnormal{d}x\textnormal{d}\widehat{e} + C,
\end{align*}
where $C$ is a constant independent of $\boldalpha$
and therefore can be safely ignored.
Let us denote the first two terms by $J(\boldalpha)$:
\begin{align}
J(\boldalpha)= J_0(\boldalpha) - C = \frac{1}{2}\boldalpha^\top \boldH \boldalpha - \boldh^\top \boldalpha,
\label{J}
\end{align}
where
\begin{align*}
  \boldH &= \iint {\bm \varphi}(x,\widehat{e}){\bm \varphi}(x,\widehat{e})^\top
  p(x)p(\widehat{e})\textnormal{d}x \textnormal{d}\widehat{e}, \\
\boldh &= \iint {\bm \varphi}(x,\widehat{e})p(x,\widehat{e})
\textnormal{d}x\textnormal{d}\widehat{e}.
\end{align*}
Approximating the expectations in $\boldH$ and $\boldh$ by empirical averages,
we obtain the following optimization problem:
\begin{align*}
\tilde{\boldalpha} = \mathop{\text{argmin}}_{\boldalpha} \Bigl[\frac{1}{2}\boldalpha^\top \widehat{\boldH}\boldalpha - \widehat{\boldh}^\top \boldalpha + \lambda \boldalpha^\top \boldalpha \Bigr], 
\end{align*}
where a regularization term $\lambda\boldalpha^\top \boldalpha$ is included for avoiding overfitting,
and
\begin{align*}
\widehat{\boldH} &= \frac{1}{n^2} \sum_{i,j = 1}^n {\bm \varphi}(x_i, \widehat{e}_j)
{\bm \varphi}(x_i, \widehat{e}_j)^\top, \nonumber \\
\widehat{\boldh} &= \frac{1}{n}\sum_{i = 1}^n {\bm \varphi}(x_i, \widehat{e}_i).
\end{align*}
Differentiating the above objective function with respect to $\boldalpha$ and equating it to zero,
we can obtain an analytic-form solution:
\begin{align}
\label{eq:LSMI_Solution}
\widehat{\boldalpha} = (\widehat{\boldH} + \lambda \boldI_b)^{-1}\widehat{\boldh},
\end{align}
where $\boldI_b$ denotes the $b$-dimensional identity matrix.
It was shown that LSMI is consistent under mild assumptions
and it achieves the optimal convergence rate
\citep{arXiv:Kanamori+etal:2009}.

Given a density ratio estimator $\widehat{r}=r_{\widehat{\boldalpha}}$,
SMI can be simply approximated as follows \citep{AISTATS:Suzuki+Sugiyama:2010}:
\begin{align}
\widehat{\textnormal{SMI}}(X, \widehat{E}) &= \frac{1}{n} \sum_{i = 1}^n \widehat{r}(x_i,\widehat{e}_i)  - \frac{1}{2n^2} \sum_{i = 1}^n \widehat{r}(x_i,\widehat{e}_i)^2 -\frac{1}{2} \nonumber \\
&=\widehat{\boldh}^\top\widehat{\boldalpha} - \frac{1}{2}\widehat{\boldalpha}^\top \widehat{\boldH} \widehat{\boldalpha}   - \frac{1}{2}. 
\label{LSMI}
\end{align}

%


\subsection{Model Selection in LSMI}
LSMI contains three tuning parameters:
the number of basis functions $b$,
the kernel width $\sigma$,
and the regularization parameter $\lambda$.
In our experiments, we fix $b = \min(200, n)$,
and choose $\sigma$ and $\lambda$ by cross-validation (CV) with grid search as follows.
First, the samples ${\cal Z}=\{z_i\;|\;z_i = (x_i, \widehat{e}_i)\}_{i=1}^n$ are divided into $K$ disjoint subsets $\{{\cal Z}_k\}_{k=1}^K$ of (approximately) the same size (we set $K = 2$ in experiments).
Then, an estimator $\widehat{\boldalpha}_{{\cal Z}_k}$ is obtained using
${\cal Z}\backslash{\cal Z}_k$ (i.e., without ${\cal Z}_k$),
and the approximation error for the hold-out samples ${\cal Z}_k$ is computed as 
\[
J_{{\cal Z}_k}^{(K \text{-CV})} = \frac{1}{2}\widehat{\boldalpha}_{{\cal Z}_k}^\top \widehat{\boldH}_{{\cal Z}_k}\widehat{\boldalpha}_{{\cal Z}_k} - \widehat{\boldh}_{{\cal Z}_k}^\top\widehat{\boldalpha}_{{\cal Z}_k},
\]
where,
for $|{\cal Z}_k|$ being the number of samples in the subset ${\cal Z}_k$, 
\begin{align*}
\widehat{\boldH}_{{\cal Z}_k} &= \frac{1}{|{\cal Z}_k|^2} \sum_{x, \widehat{e} \in {\cal Z}_k} {\bm \varphi}(x, \widehat{e})
{\bm \varphi}(x, \widehat{e})^\top, \nonumber \\
\widehat{\boldh}_{{\cal Z}_k} &= \frac{1}{|{\cal Z}_k|}\sum_{(x, \widehat{e}) \in {\cal Z}_k} {\bm \varphi}(x, \widehat{e}).
\end{align*}

This procedure is repeated for $k = 1, \ldots, K$,
and its average $J^{(K\text{-CV})}$ is outputted as
\begin{align}
  J^{(K\text{-CV})} = \frac{1}{K}\sum_{k = 1}^K J_{{\cal Z}_k}^{(K\text{-CV})}.
  \label{LSMI-CV}
\end{align}
We compute $J^{(K\text{-CV})}$ for all model candidates (the kernel width $\sigma$ and the regularization parameter $\lambda$ in the current setup), and choose the model that minimizes $J^{(K\text{-CV})}$. 
Note that $J^{(K\text{-CV})}$ is an almost unbiased estimator of the objective function \eqref{J}, where the almost-ness comes from the fact that the number of samples is reduced in the CV procedure due to data splitting \citep{book:Schoelkopf+Smola:2002}.

The LSMI algorithm is summarized in Figure~\ref{fig:LSMI}.

\begin{figure}[t]
  \centering
  \framebox{
  \begin{minipage}{0.45\linewidth}
   \begin{tabbing}
    XX\=XX\=XX\=XX\=\kill
{\bf Input:} $\{(x_i, \widehat{e}_i)\}_{i = 1}^n$, $\{\sigma_i\}_{i=1}^p$, and $\{\lambda_j\}_{j=1}^q$\\
{\bf Output:} LSMI parameter $\widehat{\boldalpha}$\\[2mm]
Compute CV score for $\{\sigma_i\}_{i=1}^p$ and $\{\lambda_j\}_{j=1}^q$ by Eq.\eqref{LSMI-CV};\\
Choose $\widehat{\sigma}$ and $\widehat{\lambda}$ that minimize the CV score;\\
Compute $\widehat{\boldalpha}$ by Eq.\eqref{eq:LSMI_Solution} with
$\widehat{\sigma}$ and $\widehat{\lambda}$;
\end{tabbing}
\end{minipage}
}
\caption{Pseudo code of the LSMI algorithm with CV.}
\label{fig:LSMI}
\end{figure}



\subsection{Least-Squares Independence Regression}

Given the SMI estimator \eqref{LSMI},
our next task is to learn the parameter $\boldbeta$
in the regression model \eqref{regression-model} as
\[
\widehat{\boldbeta}=\mathop{\text{argmin}}_{\boldbeta}
\left[ \widehat{\textnormal{SMI}}(X, \widehat{E}) + \frac{\gamma}{2} \boldbeta^\top \boldbeta  \right].
\]
We call this method {\em least-squares independence regression (LSIR)}.

For regression parameter learning, we simply employ a gradient descent method:
\begin{align}
\boldbeta \longleftarrow \boldbeta - \eta \left(\frac{\partial \widehat{\textnormal{SMI}}(X, \widehat{E})}{\partial \boldbeta}
+ \gamma \boldbeta \right),
\label{eq:LSIR_Update}
\end{align}
where $\eta$ is a step size which may be chosen in practice by some approximate line search method
such as 
\emph{Armijo's rule} \citep{Book:Patriksson:1999}.%

The partial derivative of $\widehat{\textnormal{SMI}}(X, \widehat{E})$
with respect to $\boldbeta$ can be approximately expressed as
\begin{align*}
\frac{\partial \widehat{\textnormal{SMI}}(X, \widehat{E})}{\partial \boldbeta}
\approx \sum_{l = 1}^b \widehat{\alpha}_l\frac{\partial \widehat{h}_l}{\partial \boldbeta} - \frac{1}{2}\sum_{l,l' = 1}^b \widehat{\alpha}_l \widehat{\alpha}_l' \frac{\partial \widehat{H}_{l,l'}}{\partial \boldbeta},
\end{align*}
where
\begin{align*}
\frac{\partial \widehat{h}_l}{\partial \boldbeta} &= \frac{1}{n} \sum_{i = 1}^n \frac{\partial \varphi_l(x_i, \widehat{e}_i)}{\partial \boldbeta},\\
\frac{\partial \widehat{H}_{l,l'}}{\partial \boldbeta} &= \frac{1}{n^2} \sum_{i,j = 1}^n \Biggl(\frac{\partial \varphi_l(x_i, \widehat{e}_j)}{\partial \boldbeta}\varphi_{l'}(x_j, \widehat{e}_i) + \varphi_{l}(x_i, \widehat{e}_j) \frac{\partial \varphi_l(x_j, \widehat{e}_i)}{\partial \boldbeta} \Biggr), \\
\frac{\partial \varphi_l(x,\widehat{e})}{\partial \boldbeta} &= -\frac{1}{2\sigma^2}\varphi_l(x,\widehat{e})(\widehat{e} - \widehat{v}_l){\bm \psi}(x).
\end{align*}
In the above derivation, we ignored the dependence of $\boldbeta$ on $\widehat{e}_i$.
It is possible to exactly compute the derivative in principle, but
we use this approximated expression
since it is computationally efficient.

We assumed that the mean of the noise $E$ is zero.
Taking into account this, we modify the final regressor as
\begin{align*}
  \widehat{f}(x) = f_{\widehat{\boldbeta}}(x) + \frac{1}{n}\sum_{i = 1}^n \left(y_i - f_{\widehat{\boldbeta}}(x_i)\right).
\end{align*}

\subsection{Model Selection in LSIR}
LSIR contains three tuning parameters---the number of basis functions $m$,
the kernel width $\tau$,
and the regularization parameter $\gamma$.
In our experiments, we fix $m = \min(200, n)$,
and choose $\tau$ and $\gamma$ by CV with grid search as follows.
First, the samples ${\cal Z}=\{z_i\;|\;z_i = (x_i, \widehat{e}_i)\}_{i=1}^n$ are divided into $T$ disjoint subsets $\{{\cal Z}_t\}_{t=1}^T$ of (approximately) the same size (we set $T = 2$ in experiments). Then, an estimator $\widehat{\boldbeta}_{{\cal Z}_t}$ is obtained using
${\cal Z}\backslash{\cal Z}_t$ (i.e., without ${\cal Z}_t$),
and the independence criterion for the hold-out samples ${\cal Z}_t$ is computed as
\[
\widehat{I}_{{\cal Z}_t}^{(T \text{-CV})} = \widehat{\boldh}_{{\cal Z}_t}^\top\widehat{\boldalpha}_{{\cal Z}_t} -\frac{1}{2}\widehat{\boldalpha}_{{\cal Z}_t}^\top \widehat{\boldH}_{{\cal Z}_t} \widehat{\boldalpha}_{{\cal Z}_t}- \frac{1}{2}.
\]
This procedure is repeated for $t = 1, \ldots, T$,
and its average $\widehat{I}^{(T\text{-CV})}$ is computed as
\begin{align}
  \widehat{I}^{(T\text{-CV})} = \frac{1}{T}\sum_{t = 1}^T \widehat{I}_{{\cal Z}_t}^{(T\text{-CV})}.
  \label{LSIR-CV}
\end{align}
We compute $\widehat{I}^{(T\text{-CV})}$ for all model candidates (the kernel width $\tau$ and the regularization parameter $\gamma$ in the current setup), and choose the model that minimizes $\widehat{I}^{(T\text{-CV})}$. 

The LSIR algorithm is summarized in Figure~\ref{fig:LSIR}.
A MATLAB$\textsuperscript{\textregistered}$ implementation of LSIR is available from
\begin{center}
`\texttt{http://sugiyama-www.cs.titech.ac.jp/$\tilde{~}$yamada/lsir.html}'.
\end{center}

\begin{figure}[t]
  \centering
  \framebox{
  \begin{minipage}{0.45\linewidth}
   \begin{tabbing}
    XX\=XX\=XX\=XX\=\kill
{\bf Input:} $\{(x_i, y_i)\}_{i = 1}^n$, $\{\tau_i\}_{i=1}^p$, and $\{\gamma_j\}_{j=1}^q$\\
{\bf Output:} LSIR parameter $\widehat{\boldbeta}$\\[2mm]
Compute CV score for all $\{\tau_i\}_{i=1}^p$ and $\{\gamma_j\}_{j=1}^q$
by Eq.\eqref{LSIR-CV};\\
Choose $\widehat{\tau}$ and $\widehat{\gamma}$ that minimize the CV score;\\
Compute $\widehat{\boldbeta}$ by gradient descent \eqref{eq:LSIR_Update}
with $\widehat{\tau}$ and $\widehat{\gamma}$;
\end{tabbing}
\end{minipage}
}
\caption{Pseudo code of the LSIR algorithm with CV.}
\label{fig:LSIR}
\end{figure}

%

\section{Causal Direction Inference by LSIR}
In the previous section,
we gave a dependence minimizing regression method, LSIR,
that is equipped with CV for model selection.
In this section, following \citet{NIPS:Hoyer+etal:2009},
we explain how LSIR can be used for causal direction inference.

Our final goal is, given i.i.d.~paired samples $\{(x_i, y_i)\}_{i = 1}^n$,
to determine whether $X$ causes $Y$ or vice versa.
To this end, we test whether the causal model $Y = f_Y(X) + E_Y$ 
or the alternative model $X = f_X(Y) + E_X$
fits the data well,
where the goodness of fit is measured by independence between inputs and
residuals (i.e., estimated noise).
Independence of inputs and residuals may be decided
in practice by the \emph{permutation test} \citep{book:Efron+Tibshirani:1993}.

More specifically, we first run LSIR for $\{(x_i, y_i)\}_{i = 1}^n$ as usual,
and obtain a regression function $\widehat{f}$.
This procedure also provides an SMI estimate
for $\{(x_i, \widehat{e}_i)\;|\;\widehat{e}_i=y_i-\widehat{f}(x_i)\}_{i = 1}^n$.
Next, we randomly permute the pairs of input and residual
$\{(x_i, \widehat{e}_i)\}_{i = 1}^n$ as $\{(x_i, \widehat{e}_{\kappa(i)})\}_{i = 1}^n$,
where $\kappa(\cdot)$ is a randomly generated permutation function.
Note that
the permuted pairs of samples are independent of each other
since the random permutation breaks the dependency between $X$ and $\widehat{E}$
(if exists).
Then we compute SMI estimates for the permuted data
$\{(x_i, \widehat{e}_{\kappa(i)})\}_{i = 1}^n$ by LSMI.
This random permutation process is repeated many times
(in experiments, the number of repetitions is set to $1000$),
and the distribution of SMI estimates under the null-hypothesis
(i.e., independence) is constructed.
Finally, the $p$-value is approximated by evaluating the relative ranking of the SMI estimate
computed from the original input-residual data
over the distribution of SMI estimates for randomly permuted data.

In order to decide the causal direction,
we compute the $p$-values $p_{X\rightarrow Y}$ and $p_{X\leftarrow Y}$
for both directions
$X\rightarrow Y$ (i.e., $X$ causes $Y$) and $X\leftarrow Y$ (i.e., $Y$ causes $X$).
For a given significance level $\delta$, we determine the causal direction as follows.
\begin{itemize}
\item If $p_{X\rightarrow Y}>\delta$ and $p_{X\leftarrow Y}\le\delta$,
the model $X\rightarrow Y$ is chosen.
\item If $p_{X\leftarrow Y}>\delta$ and $p_{X\rightarrow Y}\le\delta$,
the model $X\leftarrow Y$ is selected.
\item If $p_{X\rightarrow Y},p_{X\leftarrow Y}\le\delta$,
then we conclude that there is no causal relation between $X$ and $Y$.
\item If $p_{X\rightarrow Y},p_{X\leftarrow Y}>\delta$,
perhaps our modeling assumption is not correct.
\end{itemize}

When we have prior knowledge that there exists a causal relation 
between $X$ and $Y$ but their the causal direction is unknown,
we may simply compare the values of $p_{X\rightarrow Y}$ and $p_{X\leftarrow Y}$
as follows:
\begin{itemize}
\item If $p_{X\rightarrow Y}>p_{X\leftarrow Y}$, we conclude that $X$ causes $Y$.
\item Otherwise, we conclude that $Y$ causes $X$.
\end{itemize}
This simplified procedure allows us to avoid the computational expensive permutation process.

In our preliminary experiments, we empirically observed that SMI estimates obtained by LSIR tend to be affected by the basis function choice in LSIR. To mitigate this problem, we run LSIR and compute an SMI estimate $5$ times by randomly changing basis functions. Then the regression function that gives the smallest SMI estimate among $5$ repetitions is selected and the permutation test is performed for that regression function.


\section{Existing Method: HSIC Regression}
In this section, we first review the \emph{Hilbert-Schmidt independence criterion} (HSIC)
\citep{ALT:Gretton+etal:2005} and point out its potential weaknesses.
Then, we review \emph{HSIC regression} (HSICR) \citep{ICML:Mooij+etal:2009}.

\subsection{HSIC}
\label{sec:application-mutualinformation-independence-HSIC}

The \emph{Hilbert-Schmidt independence criterion} (HSIC)
\citep{ALT:Gretton+etal:2005} is a state-of-the-art measure of statistical independence
based on characteristic functions \citep[see also][]{ISR:Feuerverger:1993,thesis:Kankainen:1995}.
 Here, we review the definition of HSIC
and explain its basic properties.

Let $\calF$ be a \emph{reproducing kernel Hilbert space} (RKHS)
with reproducing kernel $K(x,x')$
\citep{AMS:Aronszajn:1950},
and
$\calG$ be another RKHS
with reproducing kernel $L(e,e')$.
Let $C$ be a \emph{cross-covariance operator} from
$\calG$ to $\calF$,
i.e., for all $f\in\calF$ and $g\in\calG$,
\begin{align*}
 \inner{f}{Cg}_{\calF}
 &=\!\iint\!\Bigg(\!\left[f(x)\!-\!\int
f(x)\density(x)\mathrm{d}x\right]\!
\left[g(e)\!-\!\int
g(e)\density(e)\mathrm{d}e\right]\!\Bigg)
 \density(x,e)\mathrm{d}x\mathrm{d}e,
\end{align*}
where $\inner{\cdot}{\cdot}_{\calF}$ denotes the inner product in $\calF$.
Thus, $C$ can be expressed as
\begin{align*}
 C
 &=\!\iint\!\Bigg(\left[K(\cdot,x)\!-\!\!\int\!
K(\cdot,x)\density(x)\mathrm{d}x\right]\!
   \otimes\!\left[L(\cdot,e)\!-\!\!\int\!
L(\cdot,e)\density(e)\mathrm{d}e\right]\!\Bigg)
   \density(x,e)\mathrm{d}x\mathrm{d}e,
\end{align*}
where `$\otimes$' denotes the \emph{tensor product},
and we used the reproducing properties:
\begin{align*}
 f(x)&=\inner{f}{K(\cdot,x)}_{\calF}
 \;\;\;\;\mbox{and}\;\;\;\;
 g(e)=\inner{g}{L(\cdot,e)}_{\calG}.
\end{align*}

\index{universal reproducing kernel Hilbert space}

The cross-covariance operator is a generalization of the
\emph{cross-covariance matrix} between random vectors.
When $\calF$ and $\calG$ are
\emph{universal RKHSs}
\citep{JMLR:Steinwart:2001} defined on compact domains
$\domainx$ and $\calE$, respectively,
the largest singular value of $C$
is zero if and only if $x$ and $e$ are independent.
Gaussian RKHSs are examples of the universal RKHS.

HSIC is defined as the the squared \emph{Hilbert-Schmidt norm}
(the sum of the squared singular values) of
the cross-covariance operator $C$:
\begin{align*}
 \mathrm{HSIC}&
 :=
 \iiiint K(x,x')L(e,e')\density(x,e)\density(x',e')
 \mathrm{d}x\mathrm{d}e\mathrm{d}x\mathrm{d}e'\\
 &\phantom{:=}
 +\left[\iint K(x,x') \density(x)\density(x')
   \mathrm{d}x\mathrm{d}x'\right]
 \left[\iint L(e,e') \density(e)\density(e')
   \mathrm{d}e\mathrm{d}e'\right]\\
 &\phantom{:=}
 -2\iint \left[\int K(x,x')\density(x')\mathrm{d}x'\right]
 \left[\int L(e,e')\density(e')\mathrm{d}e'\right]
 \density(x,e)\mathrm{d}x\mathrm{d}e.
\end{align*}

The above expression allows one to immediately obtain an empirical
estimator---with
the i.i.d.~samples $\calZ=\{(x_k,e_k)\}_{k=1}^{\nsample}$
following $\density(x,e)$,
a consistent estimator
of $\mathrm{HSIC}$
is given as
\begin{align}
\widehat{\mathrm{HSIC}}(X, E)&:=
   \frac{1}{\nsample^2}\sum_{i,i'=1}^{\nsample}K(x_i,x_{i'})L(e_i,e_{i'})
   +\frac{1}{\nsample^4}\sum_{i,i',j,j'=1}^{\nsample}K(x_i,x_{i'})L(e_j,e_{j'}) \nonumber \\
   &\phantom{:=}
   -\frac{2}{\nsample^3}\sum_{i,j,k=1}^{\nsample}K(x_i,x_k)L(e_j,e_k) \nonumber \\
   &\phantom{:}=
   \frac{1}{\nsample^2}\tr{\boldK\boldGamma\boldL\boldGamma},
\label{HSIC}
\end{align}
where
\begin{align*}
\boldK_{i,i'}&=K(x_i,x_{i'}),
\;\;\;\;
\boldL_{j,j'}=L(e_i,e_{i'}),
 \;\;\;\;\mbox{and}\;\;\;\;
\boldGamma=\boldI_{\nsample}-\frac{1}{\nsample}\boldone_{\nsample}\boldone_{\nsample}^\top.
\end{align*}
$\boldI_{\nsample}$ denotes the $\nsample$-dimensional identity matrix,
and $\boldone_{\nsample}$ denotes
the $\nsample$-dimensional vector with all ones.

$\widehat{\mathrm{HSIC}}$
depends on the choice of the universal RKHSs $\calF$ and $\calG$.
In the original HSIC paper \citep{ALT:Gretton+etal:2005},
the Gaussian RKHS with width set to the median distance between samples
was used, which is a popular heuristic in the kernel method community
\citep{book:Schoelkopf+Smola:2002}. However, to the best of our knowledge,
there is no strong theoretical justification for this heuristic.
On the other hand, the LSMI method is equipped with cross-validation,
and thus all the tuning parameters such as the Gaussian width and the
regularization parameter
can be optimized in an objective and systematic way.
This is an advantage of LSMI over HSIC.

\subsection{HSIC Regression}
In \emph{HSIC regression} (HSICR) \citep{ICML:Mooij+etal:2009}, 
the following  linear model is employed:
\begin{align}
  f_{\boldtheta}(x) = \sum_{l = 1}^n \theta_l \phi_l(x)
             = \boldtheta^\top{\bm \phi}(x),
             \label{regression-model-hsicr}
\end{align}
where $\boldtheta=(\theta_1,\ldots,\theta_n)^\top$ are regression parameters and ${\bm \phi}(x)=(\phi_1(x),\ldots,\phi_n(x))^\top$ are basis functions.
\citet{ICML:Mooij+etal:2009} proposed to use the Gaussian basis function:
\[
\phi_l(x)= \exp \left(-\frac{ (x - x_l )^2}{2\rho^2}\right),
\]
where the kernel width $\rho$ is set to the median distance between points in the samples:
\begin{align*}
  \rho &= 2^{-1/2}\textnormal{median}(\{\|x_i - x_j\|\}_{i,j=1}^n).
\end{align*}

Given the HSIC estimator \eqref{HSIC}, the parameter $\boldtheta$ in the regression model \eqref{regression-model-hsicr} is obtained by 
\begin{align}
\widehat{\boldtheta} = \mathop{\textnormal{argmin}}_{\boldtheta}\left[ \widehat{\textnormal{HSIC}}(X, Y - f_\boldtheta(X)) + \frac{\xi}{2}\boldtheta^\top \boldtheta \right],
\label{HSIC_opt}
\end{align}
where $\xi\ge0$ is the regularization parameter for avoiding overfitting.

In the HSIC estimator, the Gaussian kernels,
\begin{align*}
K(x,x') &= \exp \left(-\frac{ (x - x' )^2}{2\sigma_{\mathrm x}^2}\right)
 \;\;\;\;\mbox{and}\;\;\;\;
L(e,e') = \exp \left(-\frac{ (e - e' )^2}{2\sigma_{\mathrm e}^2}\right),
\end{align*}
are used and their kernel widths are set to the median distance between points in the samples: 
\begin{align*}
  \sigma_\mathrm{x} &= 2^{-1/2}\textnormal{median}(\{\|x_i - x_j\|\}_{i,j=1}^n),\\
  \sigma_\mathrm{e} &=  2^{-1/2}\textnormal{median}(\{\|e_i - e_j\|\}_{i,j=1}^n).
\end{align*}
The optimization problem \eqref{HSIC_opt} can be efficiently solved
by using the \emph{L-BFGS quasi-Newton method} \citep{MPB:Liu:1989} or gradient descent.

Then, the final regressor is given as
\begin{align*}
  \widehat{f}(x) = f_{\widehat{\boldtheta}}(x) + \frac{1}{n}\sum_{i = 1}^n \left(y_i - f_{\widehat{\boldtheta}}(x_i)\right).
\end{align*}
Note that, since it is not allowed to change the kernel width $\sigma_\mathrm{e}$ during the optimization \eqref{HSIC_opt}, $\sigma_\mathrm{e}$ is fixed to an estimate obtained based on an initial rough estimate of the residuals. This fact implies that, if the estimation accuracy of $\sigma_\mathrm{e}$ is poor, the overall performance of HSICR will be degraded. On the other hand, the LSIR method is equipped with cross-validation,
and thus all the tuning parameters can be optimized in an objective and systematic way. This is a significant advantage of LSIR over HSICR. 

\section{Experiments}\label{sec:experiments}
In this section, we first illustrate the behavior of LSIR using a toy example,
and then we evaluate the performance of LSIR using
benchmark datasets and real-world gene expression data.

\subsection{Illustrative Examples}
\begin{figure}[t]
  \centering
\includegraphics[width=0.6\textwidth]{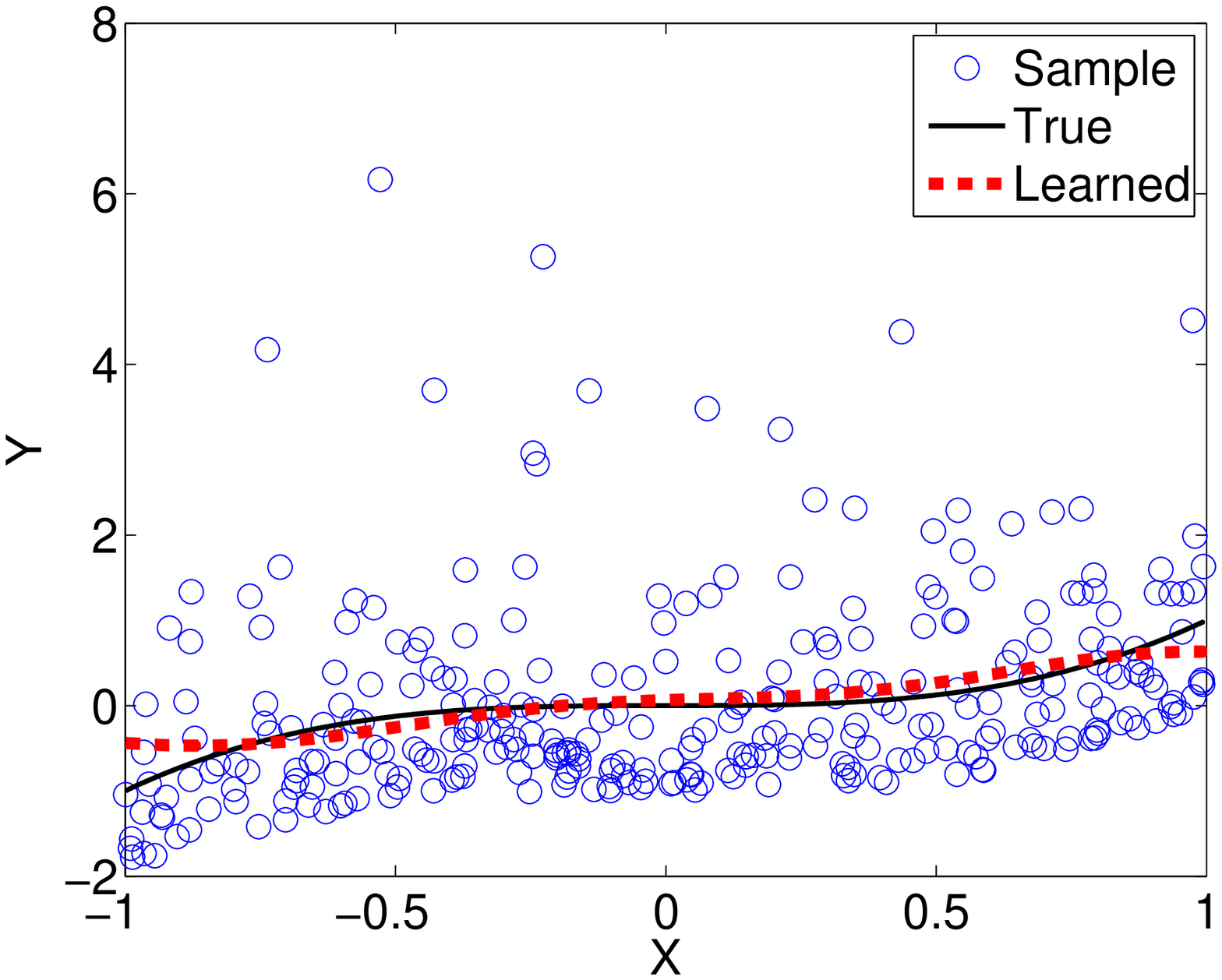}
\caption{Illustrative example.
   The solid line denotes the true function, the circles denote samples,
   and the dashed line denotes the regressor obtained by LSIR.}
 \label{fig:toy-data}
\end{figure}

Let us consider the following additive noise model: 
\[
Y = X^3 + E,
\]
where $X$ is subject to the uniform distribution on $(-1, 1)$
and $E$ is subject to the exponential distribution with rate parameter $1$
(and its mean is adjusted to have mean zero).
We drew $300$ paired samples of $X$ and $Y$
following the above generative model (see Figure~\ref{fig:toy-data}), where the ground truth is that $X$ and $E$ are independent of each other.
Thus, the null-hypothesis should be accepted (i.e., the $p$-values should be large).

\begin{figure}[t]
  \centering
\subfigure[Histogram of $p_{X\rightarrow Y}$ obtained by LSIR over $1000$ runs.
   The ground truth is to accept the null-hypothesis
   (thus the $p$-values should be large).]{
\includegraphics[width=0.3\textwidth]{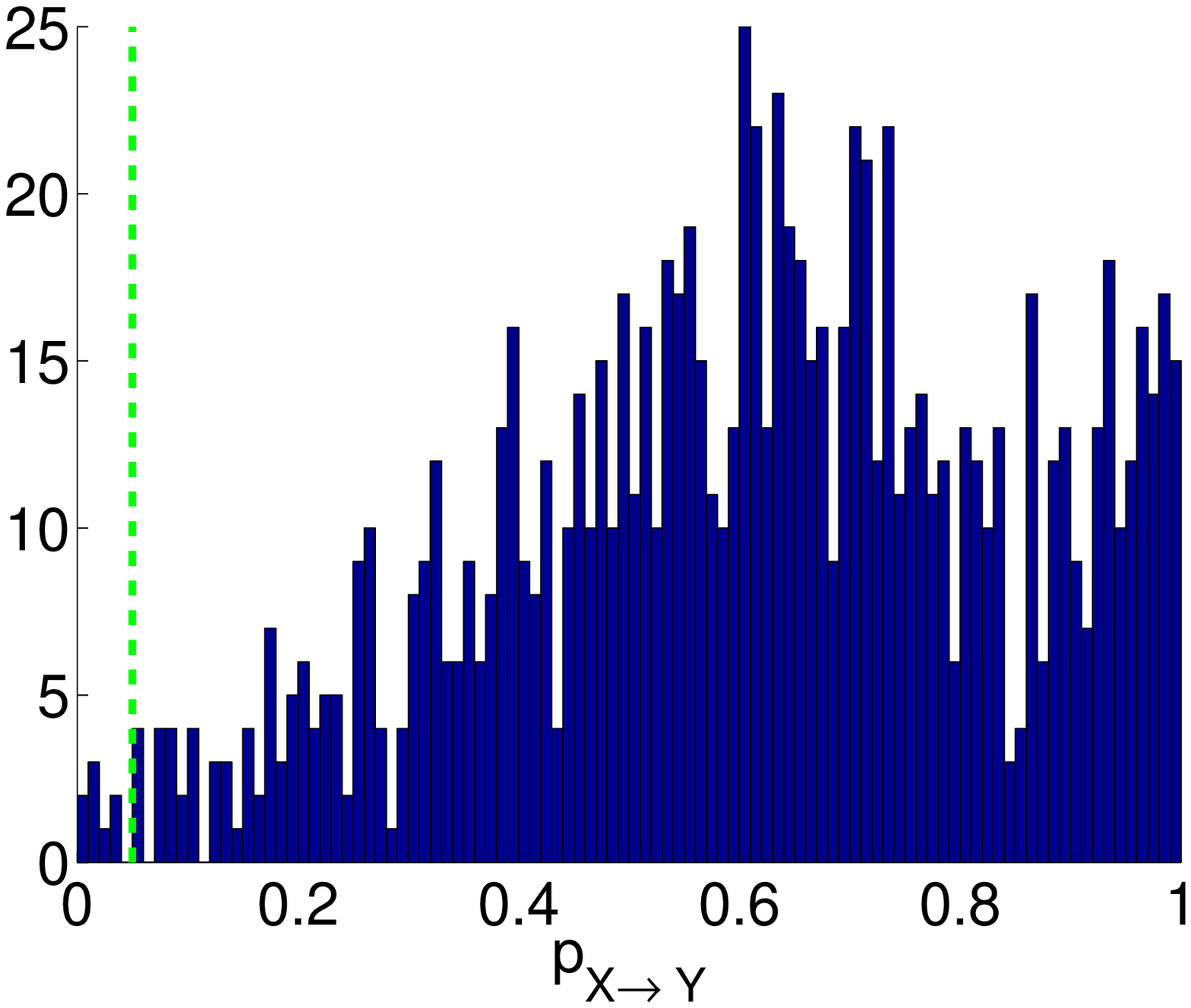}
\label{fig:toy-p-hist}
} 
~~
\subfigure[Histograms of $p_{X\leftarrow Y}$ obtained by LSIR over $1000$ runs.
   The ground truth is to reject the null-hypothesis
   (thus the $p$-values should be small).]{
\includegraphics[width=0.3\textwidth]{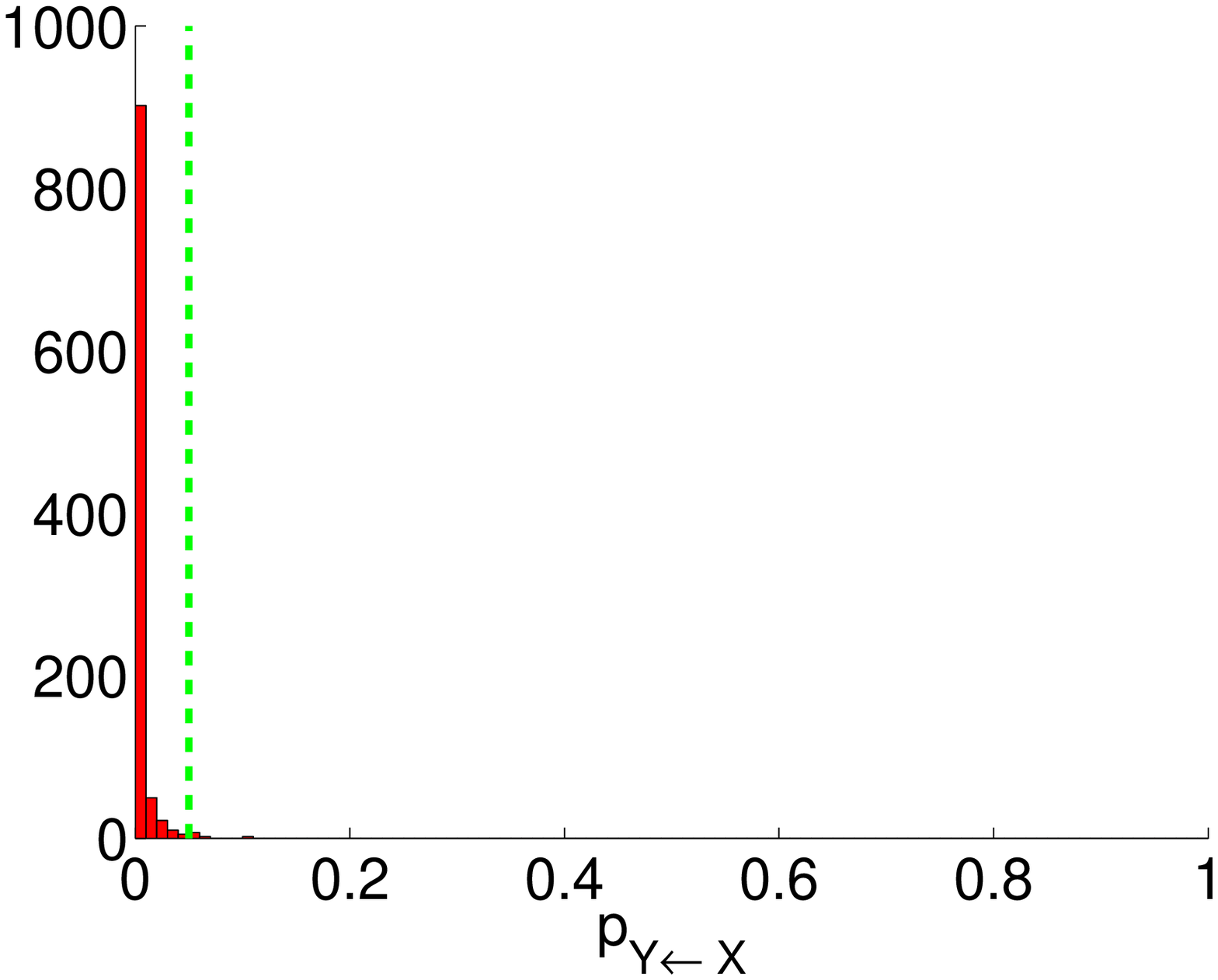}
\label{fig:toy-n-hist}
} \\
  \subfigure[Comparison of $p$-values for both directions ($p_{X\rightarrow Y}$ vs.~$p_{X\leftarrow Y}$).
  Points below the diagonal line indicate successful trials.]{
 \includegraphics[width=0.3\textwidth]{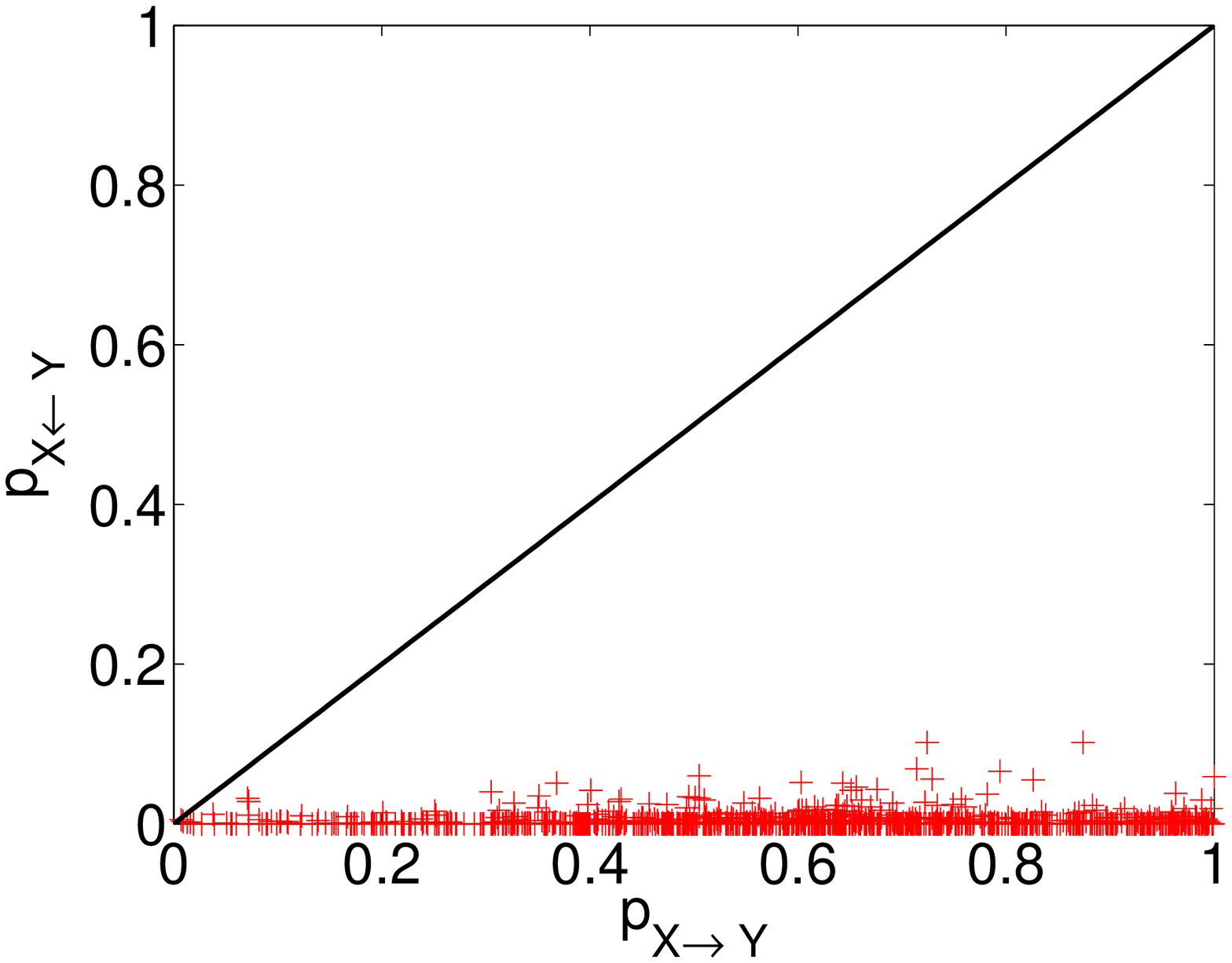}
    \label{fig:toy-p-vs}
  }~~
  \subfigure[Comparison of values of independence measures for both directions
   ($\widehat{\textnormal{SMI}}_{X\rightarrow Y}$ vs.~$\widehat{\textnormal{SMI}}_{X\leftarrow Y}$).
   Points above the diagonal line indicate successful trials.]{
 \includegraphics[width=0.3\textwidth]{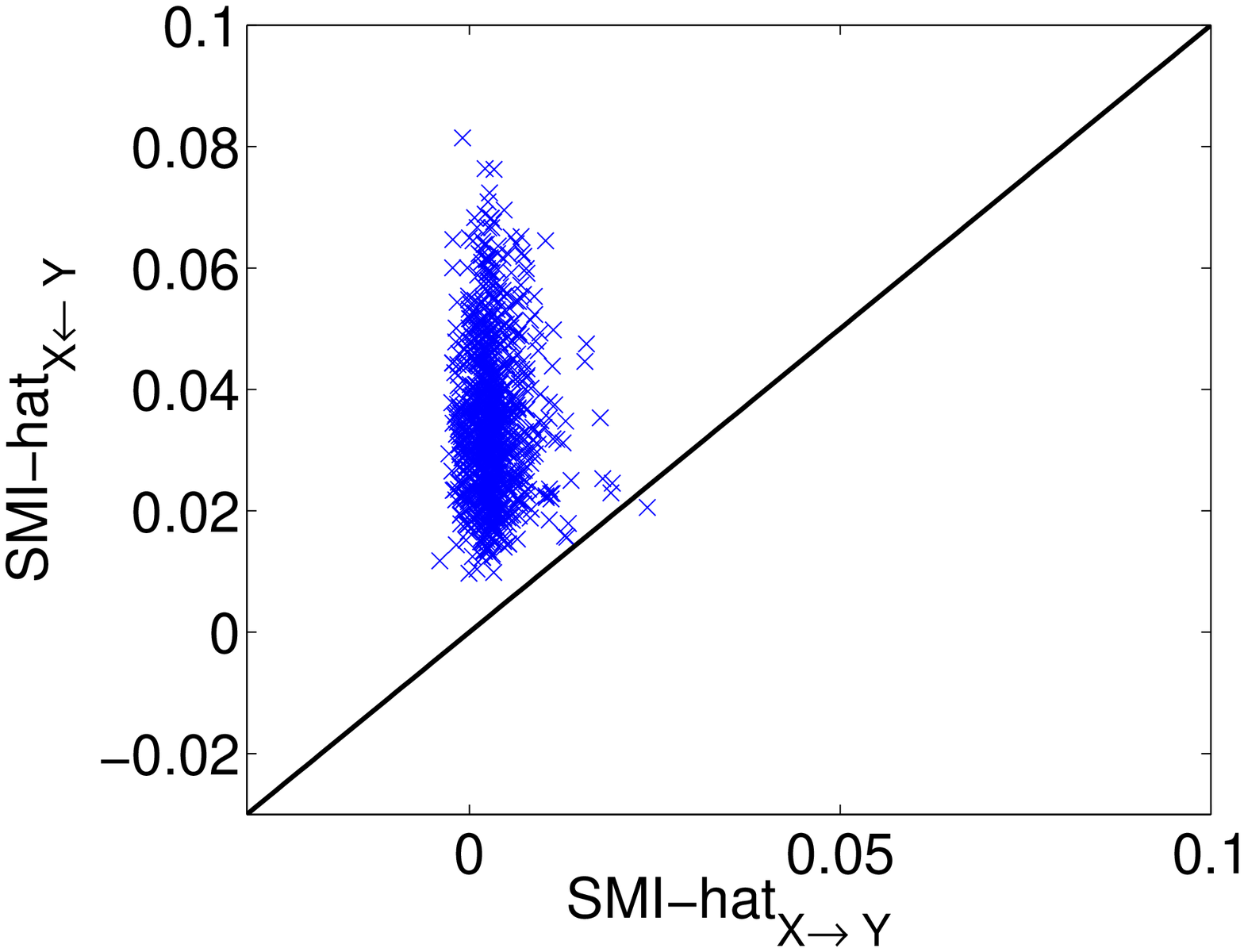}
    \label{fig:toy-measure-vs}
}
 \caption{LSIR performance statistics in illustrative example.
   }
    \label{fig:toy}
\end{figure}

Figure~\ref{fig:toy-data} depicts the regressor obtained by LSIR,
giving a good approximation to the true function.
We repeated the experiment $1000$ times with the random seed changed.
For the significance level $5\%$,
LSIR successfully accepted the null-hypothesis $992$ times out of $1000$ runs.

As \citet{ICML:Mooij+etal:2009} pointed out,
beyond the fact that the $p$-values frequently exceed the pre-specified significance level,
it is important to have a wide margin beyond the significance level
in order to cope with, e.g., multiple variable cases.
Figure~\ref{fig:toy-p-hist} depicts the histogram of $p_{X\rightarrow Y}$ obtained by LSIR
over $1000$ runs.
The plot shows that LSIR tends to produce much larger $p$-values than the significance level;
the mean and standard deviation of the $p$-values over $1000$ runs 
are $0.6114$ and $0.2327$, respectively.

Next, we consider the backward case
where the roles of $X$ and $Y$ were swapped.
In this case, the ground truth is that the input and the residual are dependent
(see Figure~\ref{fig:toy-data}).
Therefore, the null-hypothesis should be rejected
(i.e., the $p$-values should be small).
Figure~\ref{fig:toy-n-hist}
shows the histogram of $p_{X\leftarrow Y}$ obtained by LSIR
over $1000$ runs.
LSIR rejected the null-hypothesis $989$ times out of $1000$ runs;
the mean and standard deviation of the $p$-values over $1000$ runs 
are $0.0035$ and $0.0094$, respectively.

Figure~\ref{fig:toy-p-vs} depicts
the $p$-values for both directions in a trial-wise manner.
The graph shows that LSIR perfectly estimates the correct causal direction (i.e., $p_{X\rightarrow Y}> p_{X\leftarrow Y}$), 
and the \emph{margin} between $p_{X\rightarrow Y}$ and $p_{X\leftarrow Y}$ seems to be clear
(i.e., most of the points are clearly below the diagonal line).
This illustrates the usefulness of LSIR
in causal direction inference.

Finally, we investigate the values of independence measure $\widehat{\textnormal{SMI}}$,
which are plotted in Figure~\ref{fig:toy-measure-vs}
again in a trial-wise manner.
The graph implies that the values of $\widehat{\textnormal{SMI}}$ may be simply used 
for determining the causal direction, instead of the $p$-values.
Indeed, the correct causal direction (i.e., $\widehat{\textnormal{SMI}}_{X\rightarrow Y}< \widehat{\textnormal{SMI}}_{X\leftarrow Y}$)
can be found $999$ times out of $1000$ trials by this simplified method.
This would be a practically useful heuristic since
we can avoid performing the computationally intensive permutation test.

\begin{figure}[t]
  \centering
  \subfigure[dataset1]{
 \includegraphics[width=0.23\textwidth]{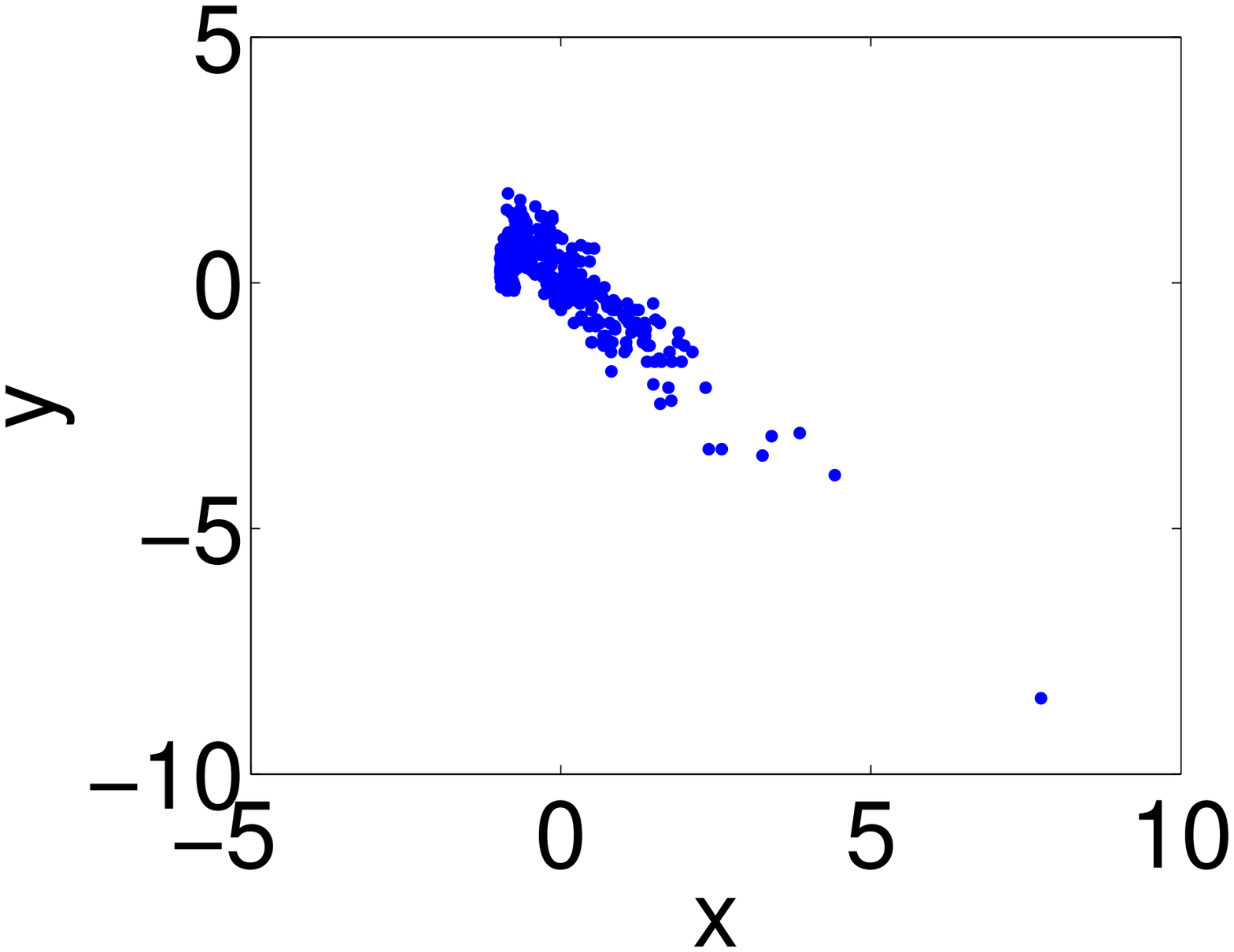}
      }
  \subfigure[dataset2]{
 \includegraphics[width=0.23\textwidth]{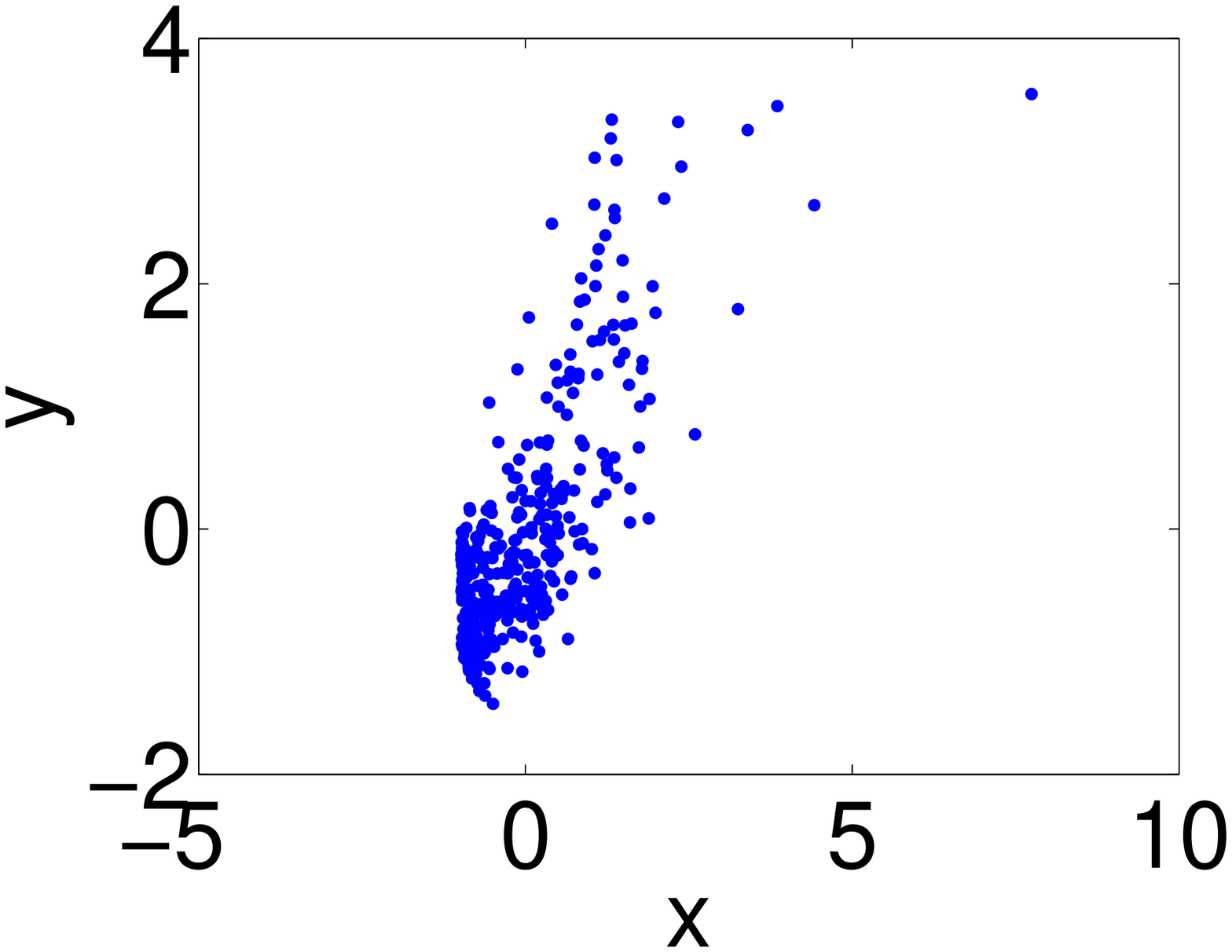}
    }
  \subfigure[dataset3]{
 \includegraphics[width=0.23\textwidth]{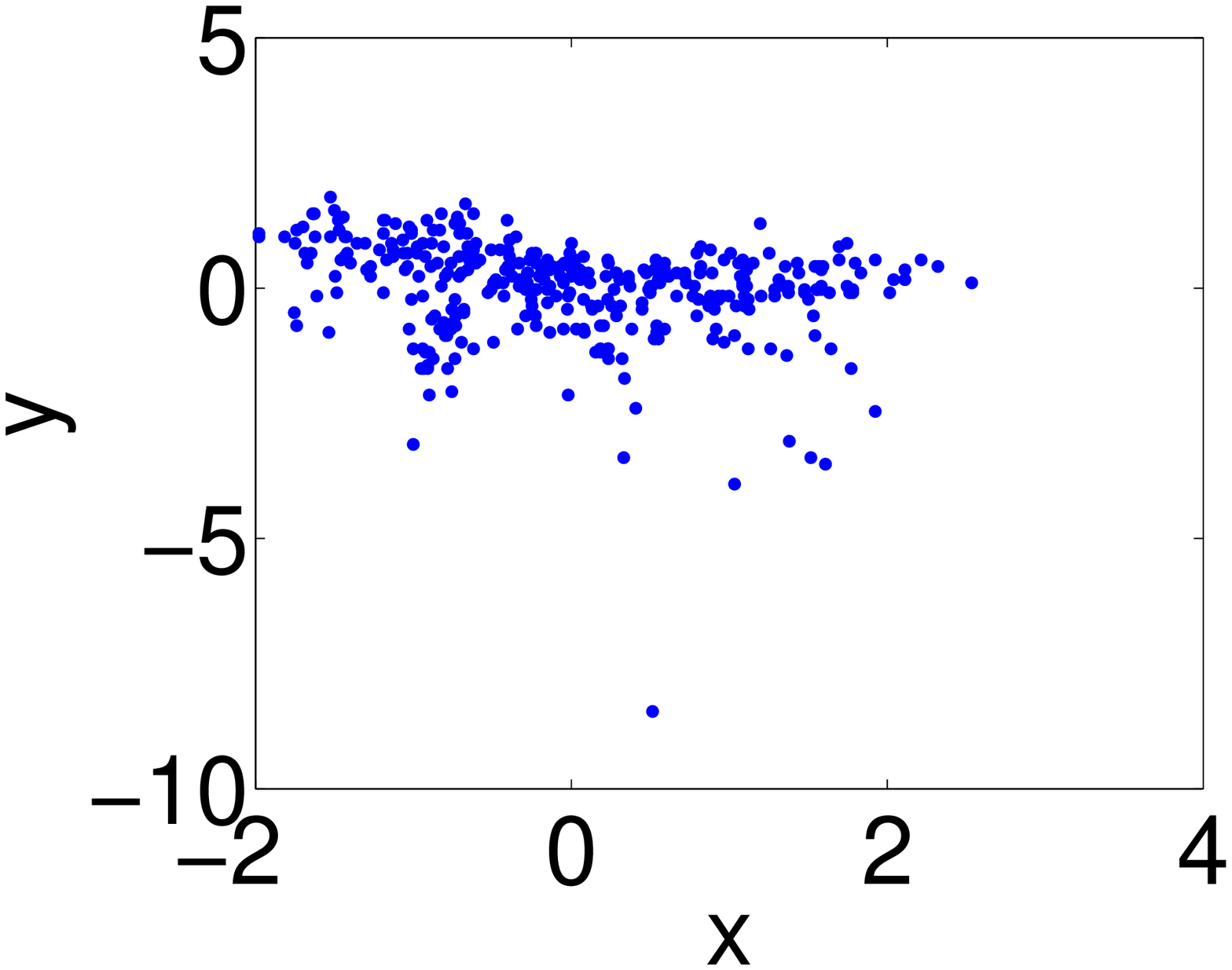}
      }
  \subfigure[dataset4]{
 \includegraphics[width=0.23\textwidth]{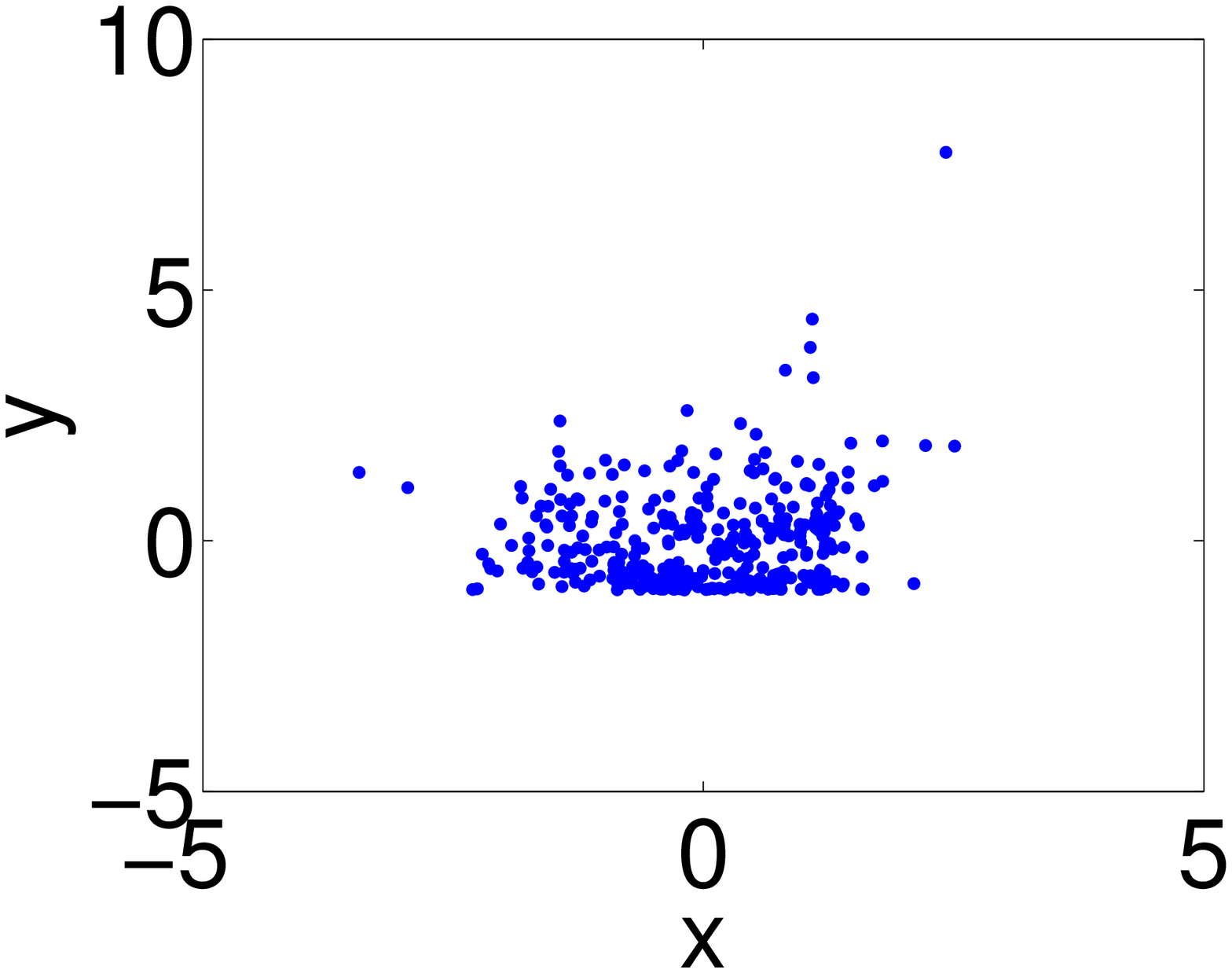}
    }
  \subfigure[dataset5]{
 \includegraphics[width=0.23\textwidth]{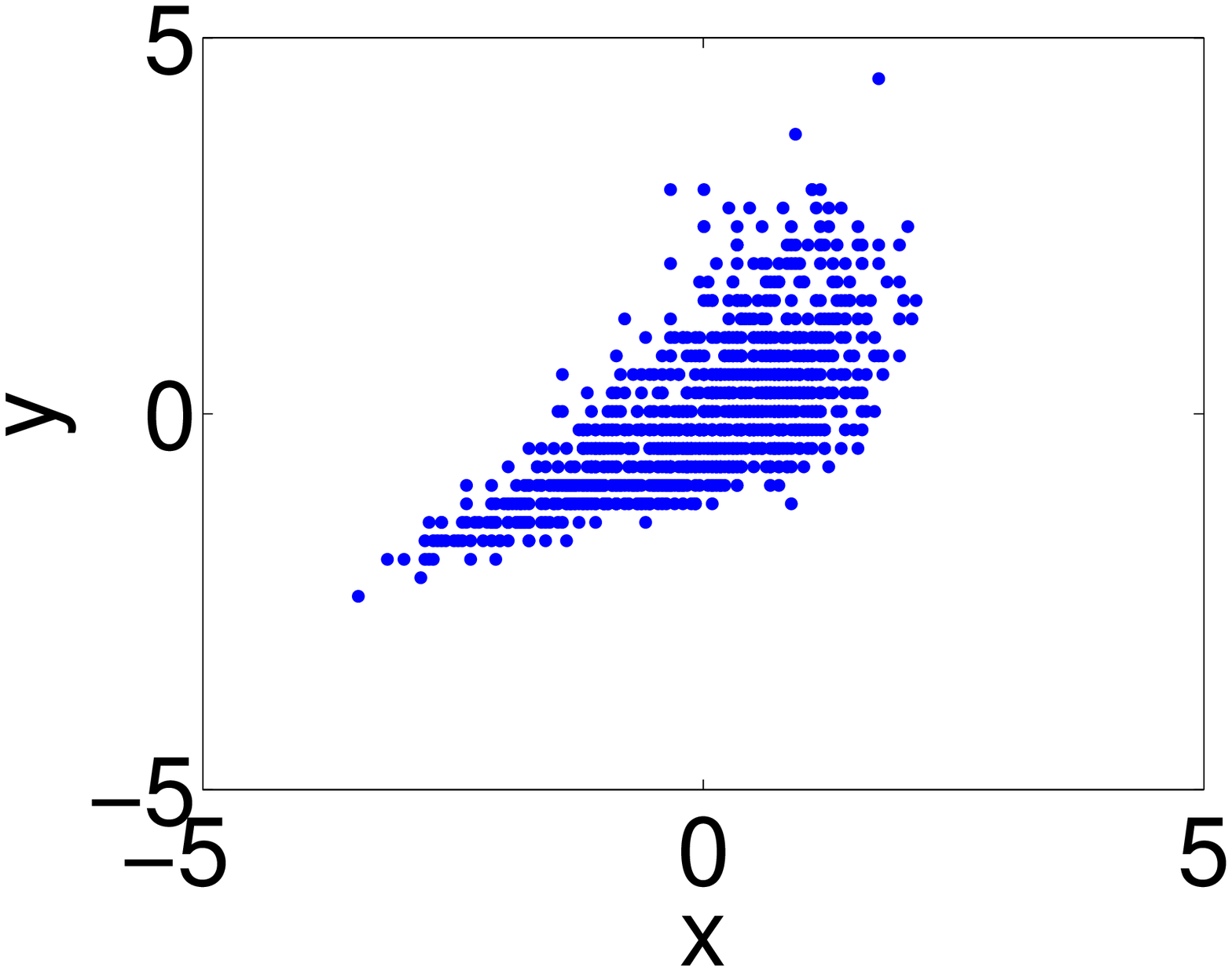}
      }
  \subfigure[dataset6]{
 \includegraphics[width=0.23\textwidth]{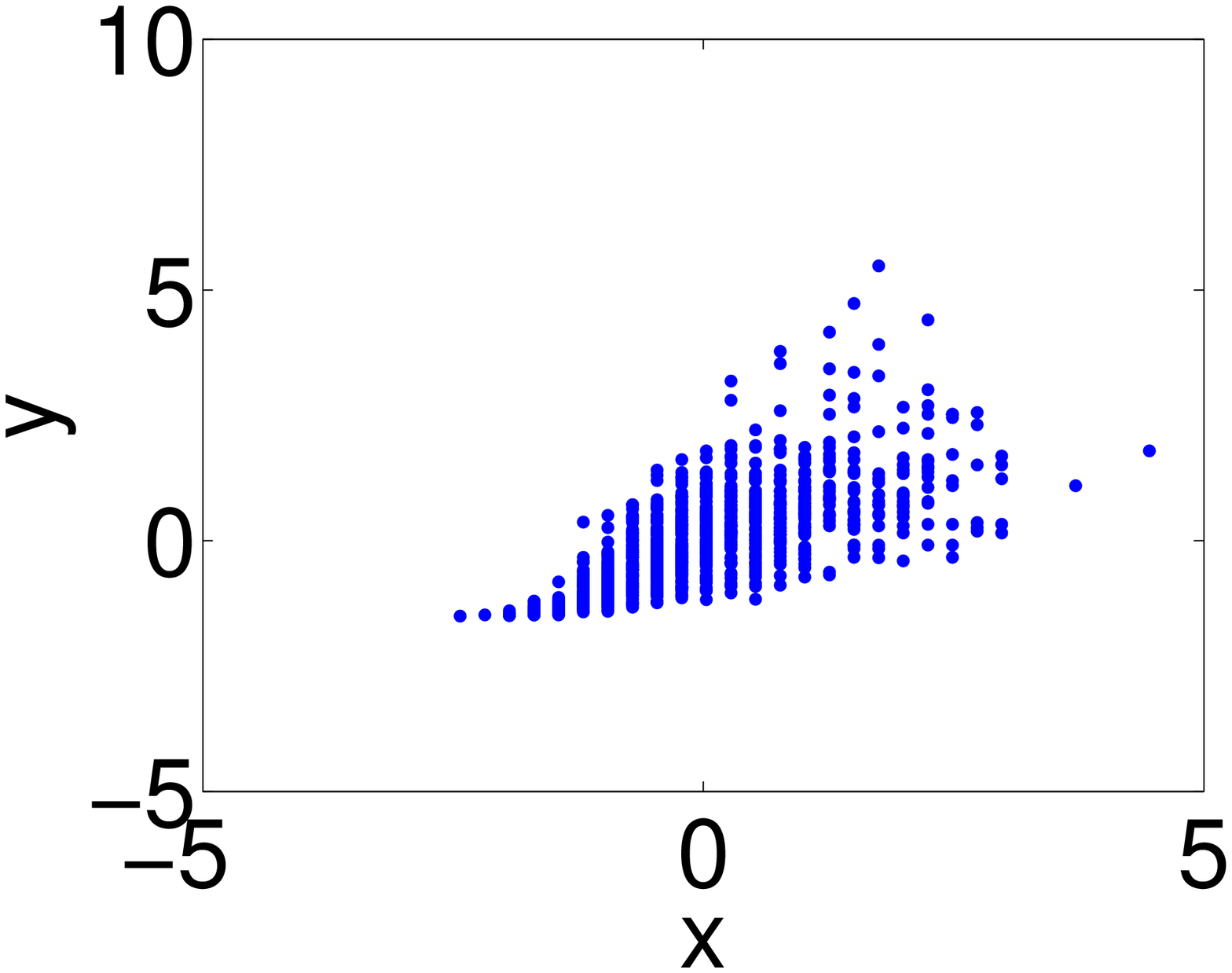}
    }
  \subfigure[dataset7]{
 \includegraphics[width=0.23\textwidth]{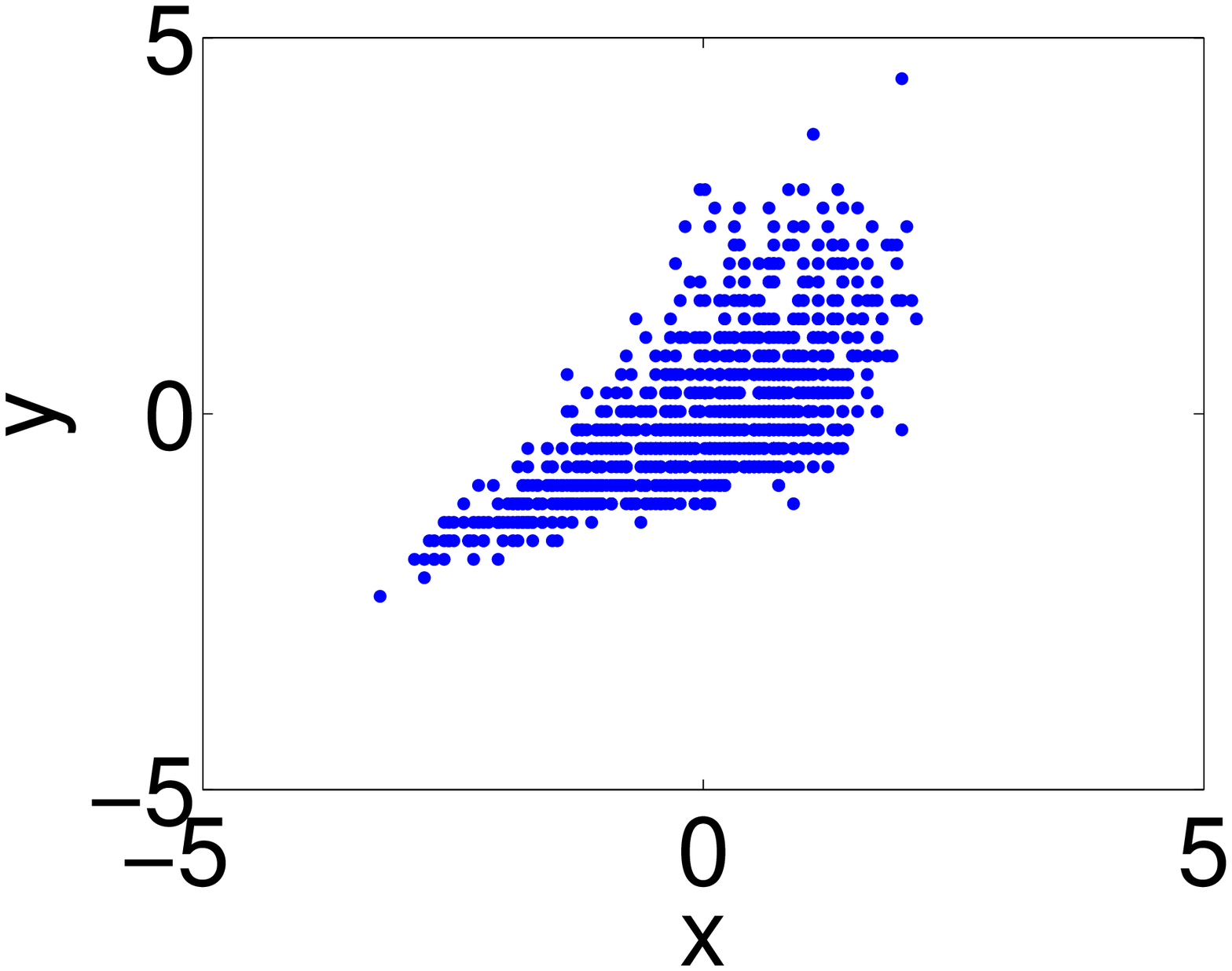}
      }
  \subfigure[dataset8]{
 \includegraphics[width=0.23\textwidth]{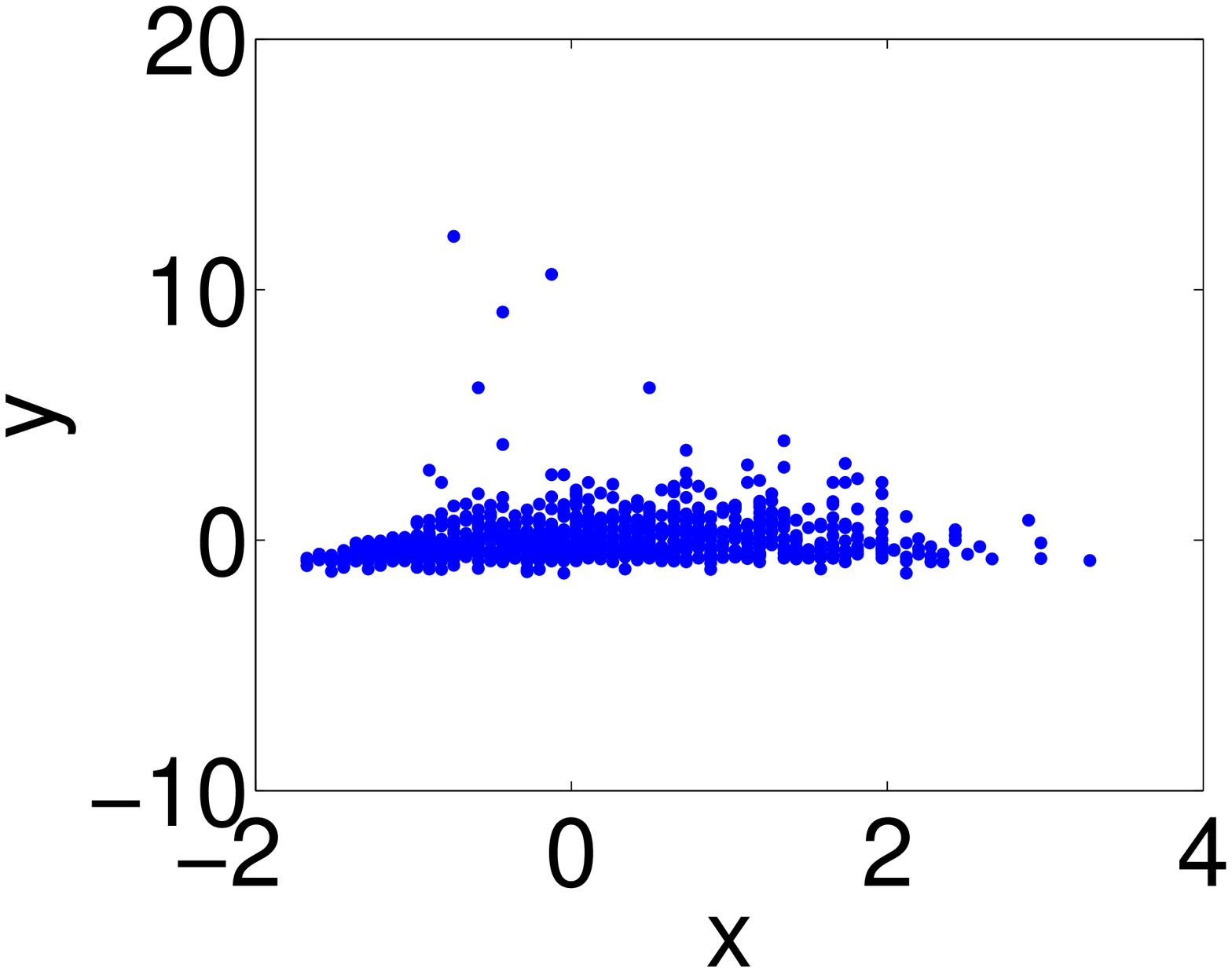}
    }
 \caption{Datasets of the `\emph{Cause-Effect Pairs}' task in
  the \emph{NIPS 2008 Causality Competition} \citep{NIPS:Mooij:2008}.}
    \label{fig:benchmark}
\end{figure}

\subsection{Benchmark Datasets}
Next, we evaluate the performance of LSIR
on the `{\em Cause-Effect Pairs}' task
in the \emph{NIPS 2008 Causality Competition} \citep{NIPS:Mooij:2008}.
The task contains $8$ datasets  (see Figure~\ref{fig:benchmark}), each has two statistically dependent random variables possessing inherent causal relationship.  The goal is to identify the causal direction from the observational data. Since these datasets consist of real-world samples, our modeling assumption may be only approximately satisfied. Thus, identifying causal directions in these datasets would be highly challenging.

The $p$-values and the independence scores for each dataset and each direction
are summarized in Table~\ref{tb:NIPS_causal}. 
 The values of HSICR, which were also computed by the permutation test, were taken from
\citep{ICML:Mooij+etal:2009},
but the $p$-values were rounded off to three decimal places
to be consistent with the results of LSIR.
When the $p$-values of both directions are less than $10^{-3}$,
we concluded that the causal direction cannot be determined (indicated by `?').

Table~\ref{tb:NIPS_causal} shows that
LSIR successfully found the correct causal direction for $7$ out of $8$ cases, while HSICR gave the correct decision only for $5$ out of $8$ cases. This implies that LSIR compares favorably with HSICR.

The values of independence measures described in Table~\ref{tb:NIPS_causal}
show that merely comparing the values of $\widehat{\textnormal{SMI}}$
is again sufficient for deciding the correct causal direction in LSIR
(see the estimated causal directions described in the brackets).
Actually, this heuristic also allows us to correctly identify the causal direction in Dataset 8.
On the other hand, in HSICR, this convenient heuristic is not 
as useful as in the case of LSIR.


\begin{table}[t]
\centering
\caption{Results for the `\emph{Cause-Effect Pairs}' task in
  the \emph{NIPS 2008 Causality Competition} \citep{NIPS:Mooij:2008}.
 When the $p$-values of both directions are less than $10^{-3}$,
 we concluded that the causal direction cannot be determined (indicated by `?').
 Estimated directions in the brackets are determined
 based on comparing the values
 of $\widehat{\textnormal{SMI}}$ or $\widehat{\textnormal{HSIC}}$.
}
\vspace*{2mm}
\label{tb:NIPS_causal}
{
(a) LSIR\\
\begin{tabular}{|c||c|c|c|c|cc|c|}
\hline
Dataset &\multicolumn{2}{c|}{$p$-values}  &
 \multicolumn{2}{c|}{$\widehat{\textnormal{SMI}}$}  &
\multicolumn{3}{c|}{Direction}\\ \cline{2-8}
 & $X\rightarrow Y$ & $X \leftarrow Y$ & $X\rightarrow Y$ & $X \leftarrow Y$ 
& \multicolumn{2}{c|}{Estimated} & Truth\\
\hline\hline
1& 0.031   & $< 10^{-3}$     &  0.0057 &  0.0265 & $\rightarrow$ & ($\rightarrow$) & $\rightarrow$\\ 
2& 0.004   & $< 10^{-3}$     &  0.0182 &  0.0301 & $\rightarrow$ & ($\rightarrow$) &$\rightarrow$ \\ 
3& 0.099   & 0.009           &  0.0090 &  0.0147 & $\rightarrow$ & ($\rightarrow$) & $\rightarrow$\\ 
4& 0.102   & 0.173           &  0.0075 &  0.0051 & $\leftarrow$ & ($\leftarrow$) & $\leftarrow$\\ 
5& $< 10^{-3}$ & 0.012           &  0.0234 &  0.0108 & $\leftarrow$ & ($\leftarrow$) & $\leftarrow$\\ 
6& 0.058 &   0.001 &  0.0079 &  0.0154 & $\rightarrow$ & ($\rightarrow$) &$\rightarrow$ \\ 
7& 0.009 & 0.018       &  0.0121 &  0.0110 & $\leftarrow$ & ($\leftarrow$)& $\leftarrow$\\ 
8& $< 10^{-3}$ & $< 10^{-3} $&  0.0149 &  0.0244 & ? & ($\rightarrow$) & $\rightarrow$\\ \hline
\end{tabular} \\
\vspace{0.3cm}
(b) HSICR\\
\begin{tabular}{|c||c|c|c|c|cc|c|}
\hline
Dataset &\multicolumn{2}{c|}{$p$-values}  &
 \multicolumn{2}{c|}{$\widehat{\textnormal{HSIC}}$} &
\multicolumn{3}{c|}{Direction}\\ \cline{2-8}
 & $X\rightarrow Y$ & $X \leftarrow Y$ & $X\rightarrow Y$ & $X \leftarrow Y$ 
& \multicolumn{2}{c|}{Estimated} & Truth\\
\hline\hline
1& 0.290 & $< 10^{-3} $& 0.0012& 0.0060& $\rightarrow$& ($\rightarrow$) & $\rightarrow$\\ 
2& 0.037 & 0.014 & 0.0020& 0.0021& $\rightarrow$& ($\rightarrow$) & $\rightarrow$\\ 
3& 0.045 & 0.003 & 0.0019 & 0.0026& $\rightarrow$& ($\rightarrow$) & $\rightarrow$\\ 
4& 0.376 & 0.012 & 0.0011& 0.0023& $\rightarrow$& ($\rightarrow$) & $\leftarrow$\\ 
5& $< 10^{-3} $& 0.160 & 0.0028& 0.0005& $\leftarrow$& ($\leftarrow$)& $\leftarrow$\\ 
6& $< 10^{-3} $& $< 10^{-3} $& 0.0032& 0.0026& ? & ($\leftarrow$)& $\rightarrow$\\ 
7& $< 10^{-3} $& 0.272 & 0.0021& 0.0005& $\leftarrow$& ($\leftarrow$)& $\leftarrow$\\ 
8& $< 10^{-3} $ & $< 10^{-3} $& 0.0015& 0.0017& $?$ & ($\rightarrow$) & $\rightarrow$\\ \hline
\end{tabular}
}
\end{table}

\begin{figure}[t]
  \centering
  \subfigure[lexA vs.~uvrA]{
  \includegraphics[width=0.23\textwidth]{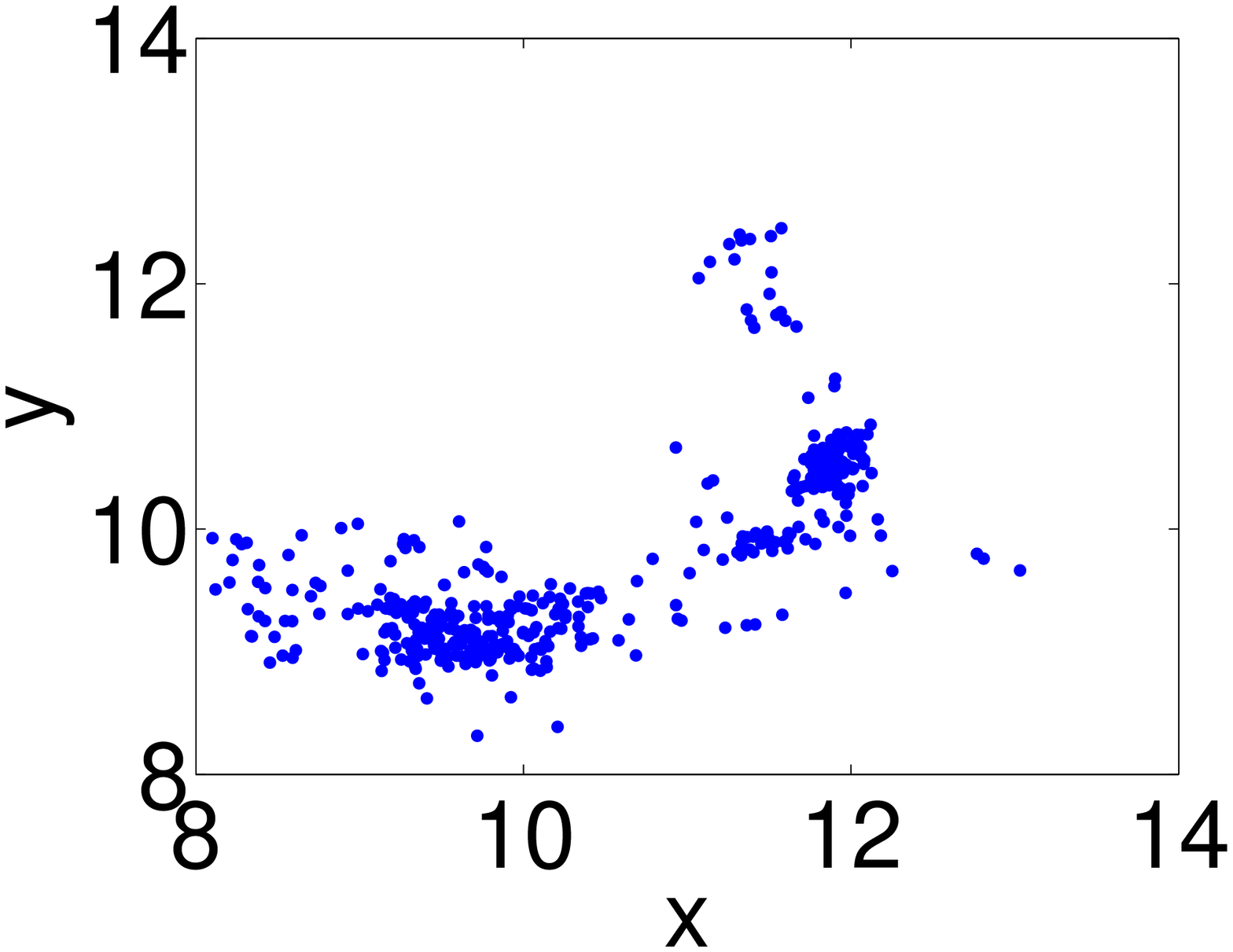}
      }
  \subfigure[lexA vs.~recA]{
    \includegraphics[width=0.23\textwidth]{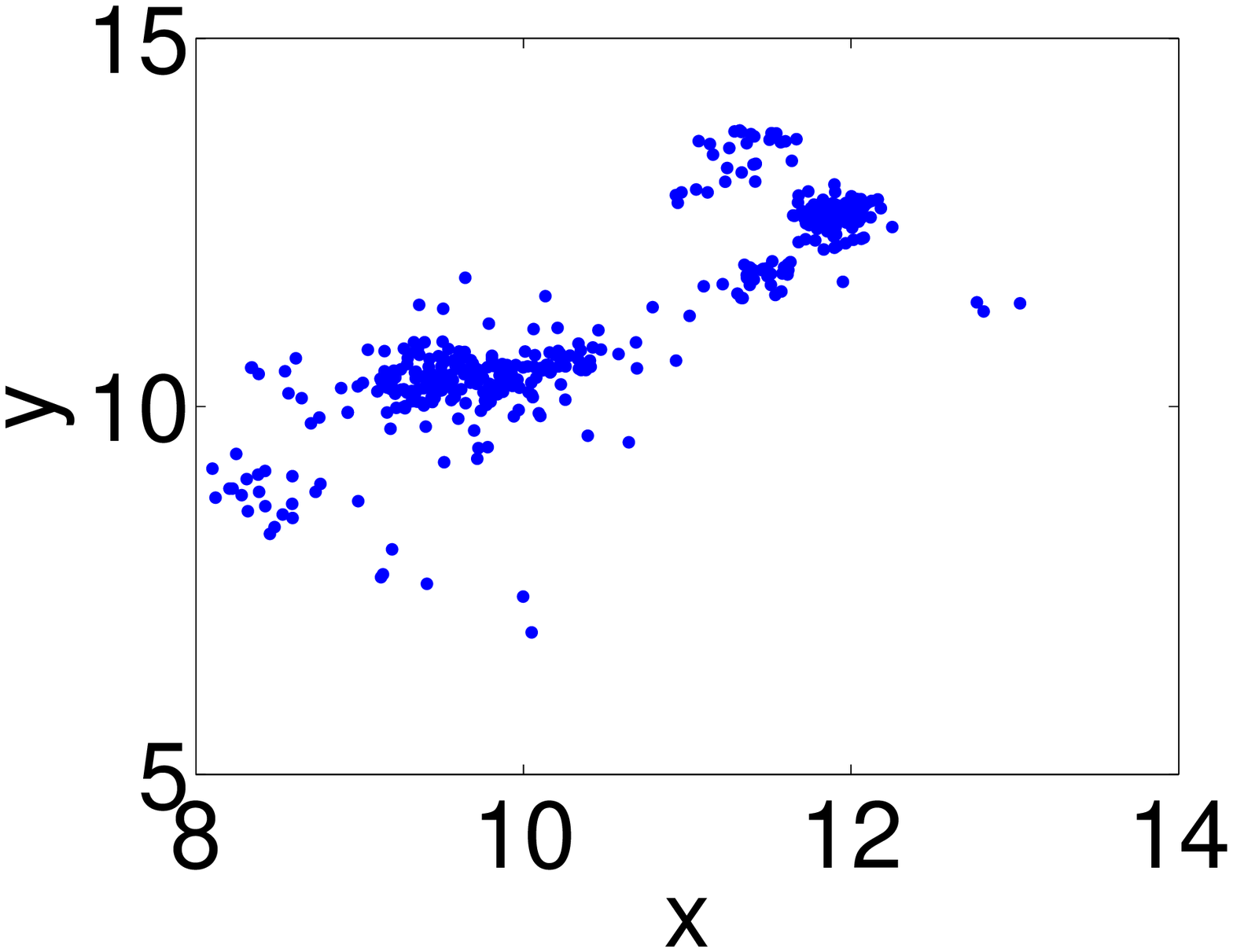}
    }
  \subfigure[lexA vs.~uvrB]{
   \includegraphics[width=0.23\textwidth]{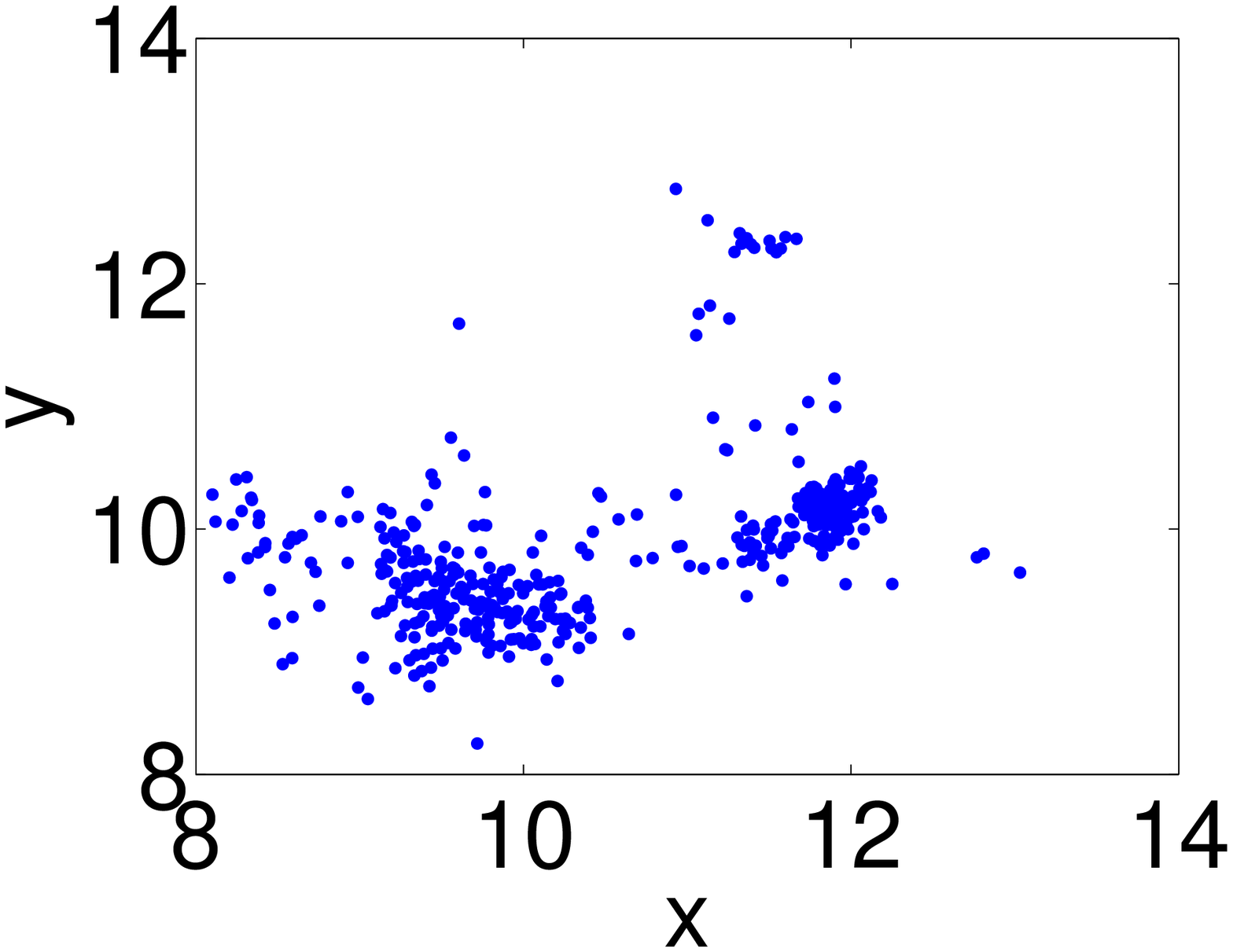}
      }
  \subfigure[lexA vs.~uvrD]{
   \includegraphics[width=0.23\textwidth]{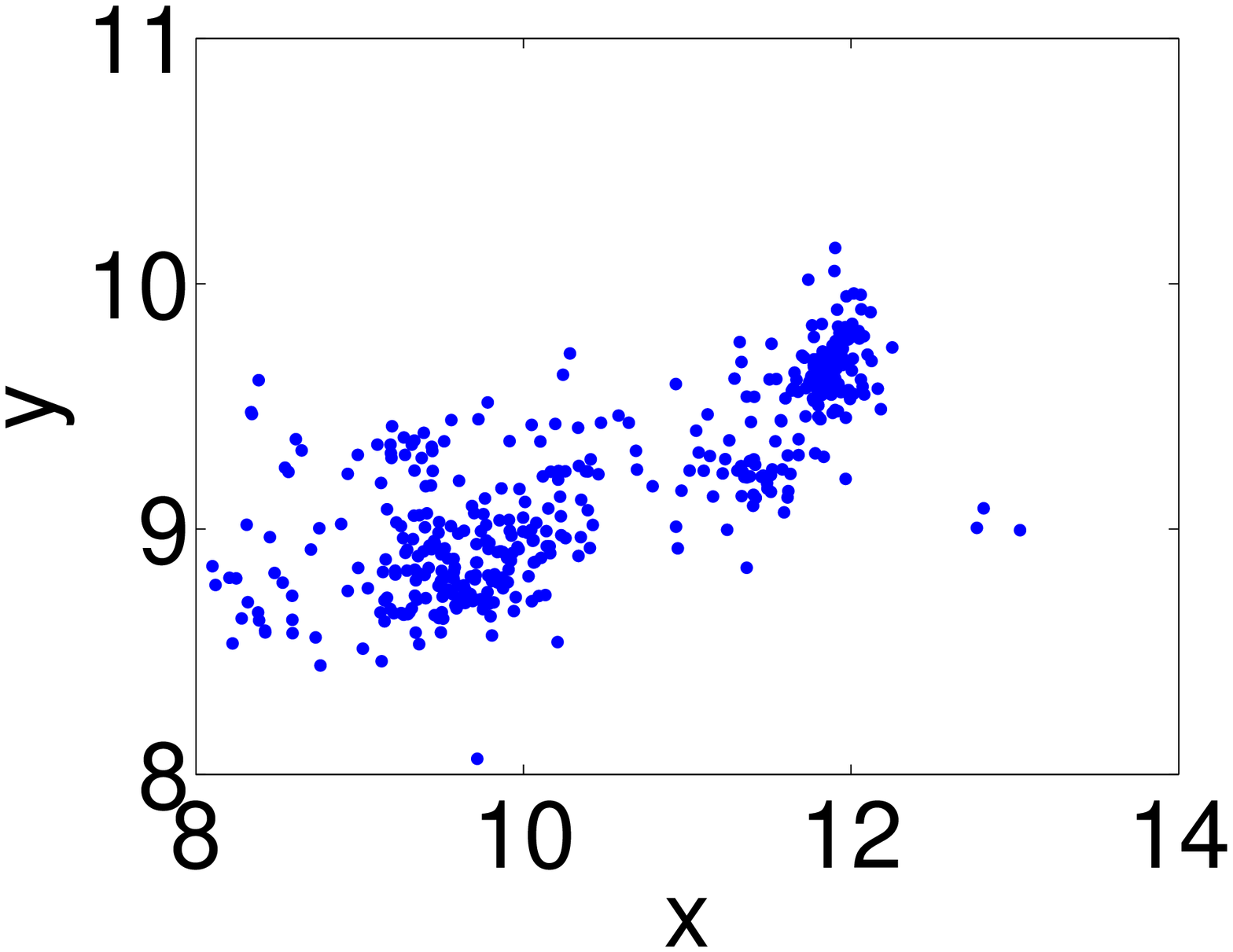}
      }
  \subfigure[crp vs.~lacA]{
   \includegraphics[width=0.23\textwidth]{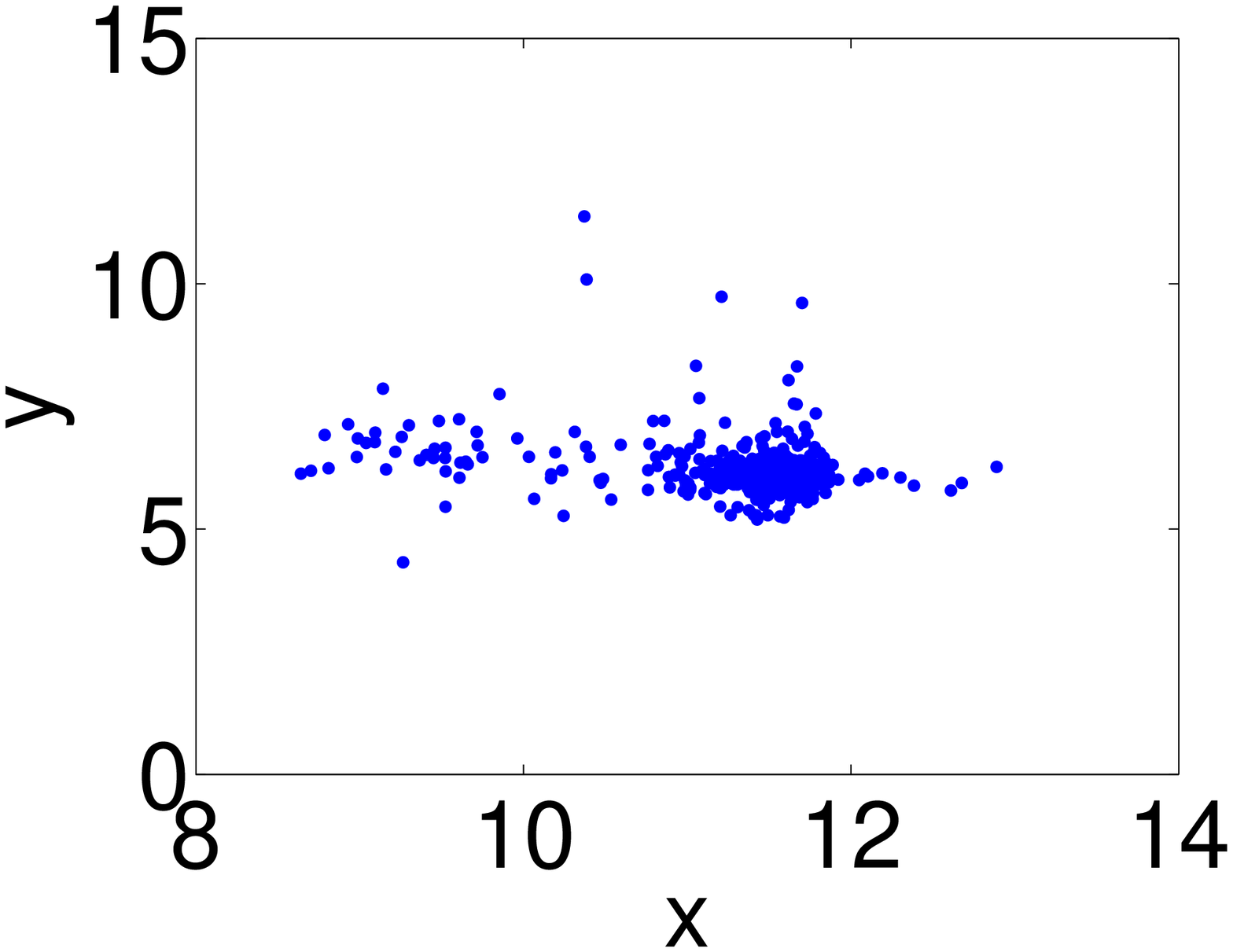}
    }
  \subfigure[crp vs.~lacY]{
  \includegraphics[width=0.23\textwidth]{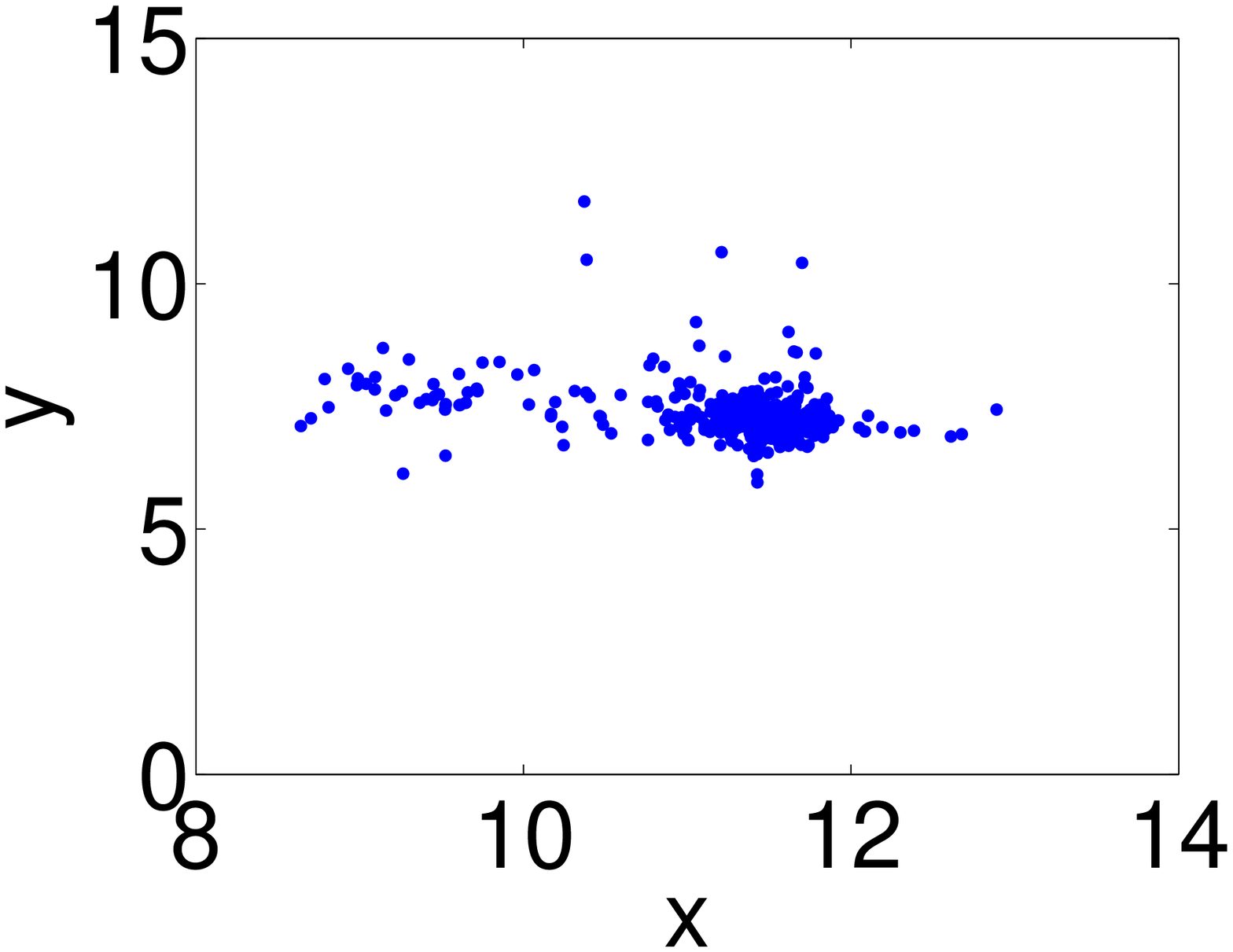}
      }
  \subfigure[crp vs.~lacZ]{
   \includegraphics[width=0.23\textwidth]{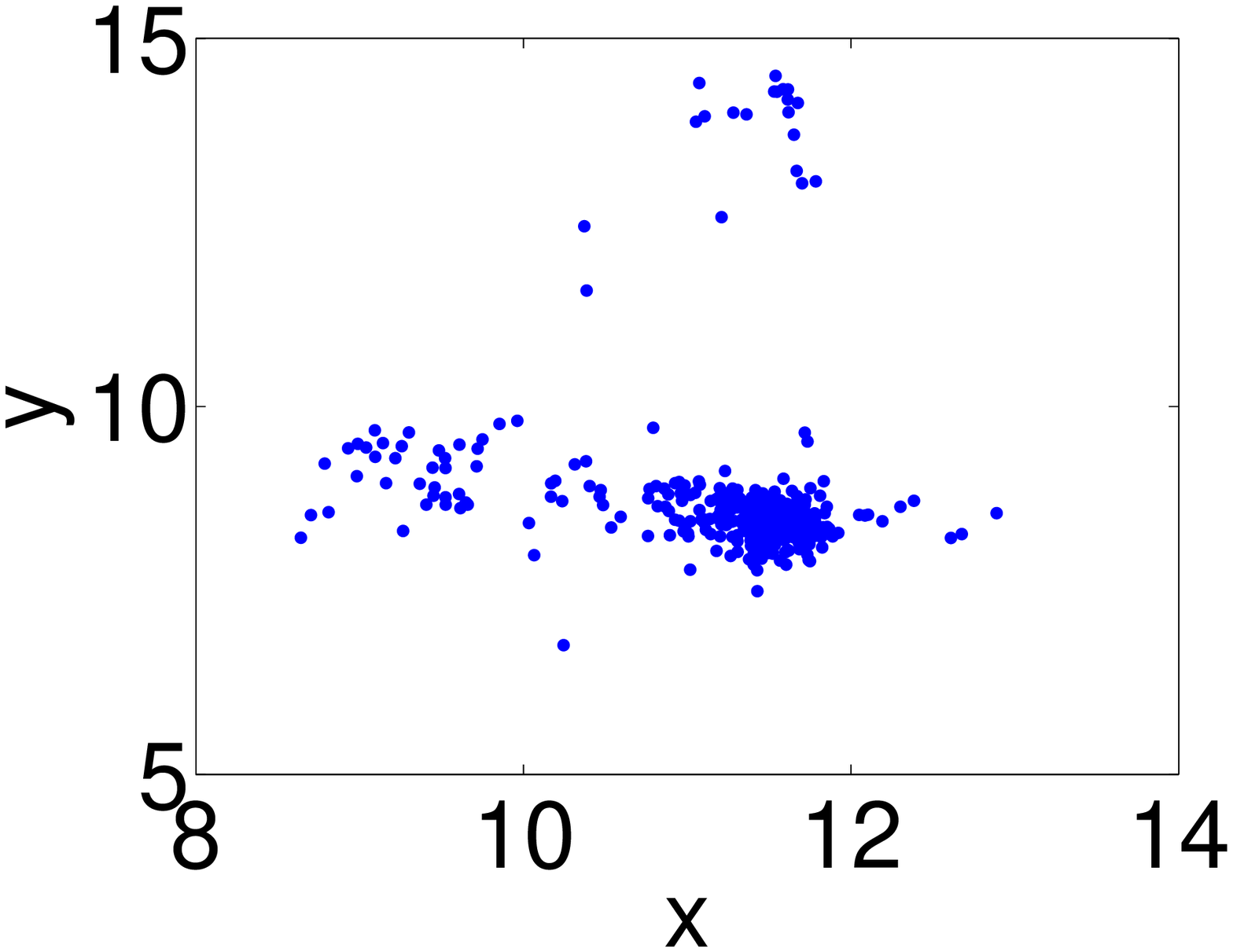}
    }
  \subfigure[lacI vs.~lacA]{
   \includegraphics[width=0.23\textwidth]{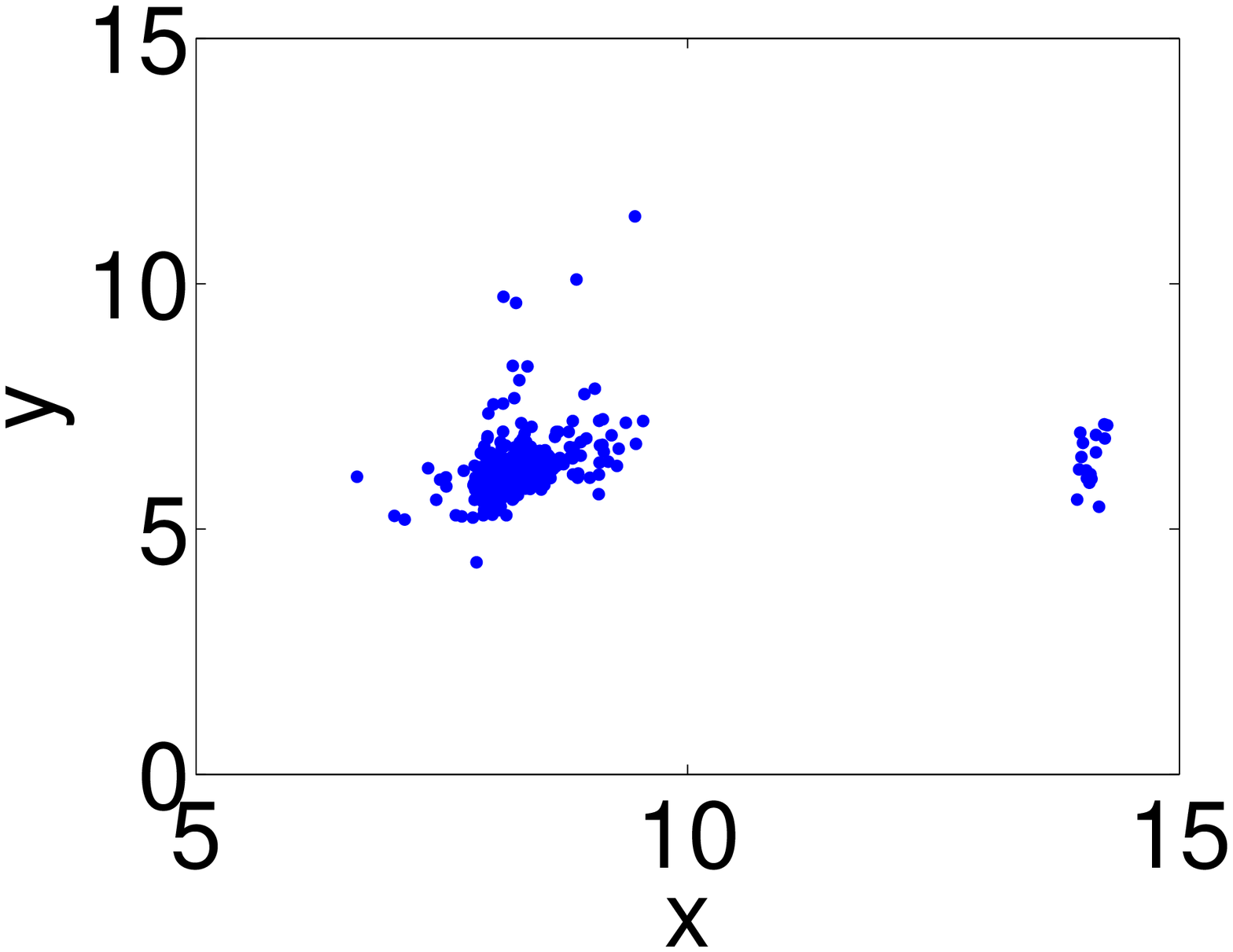}
      }
  \subfigure[lacI vs.~lacZ]{
    \includegraphics[width=0.23\textwidth]{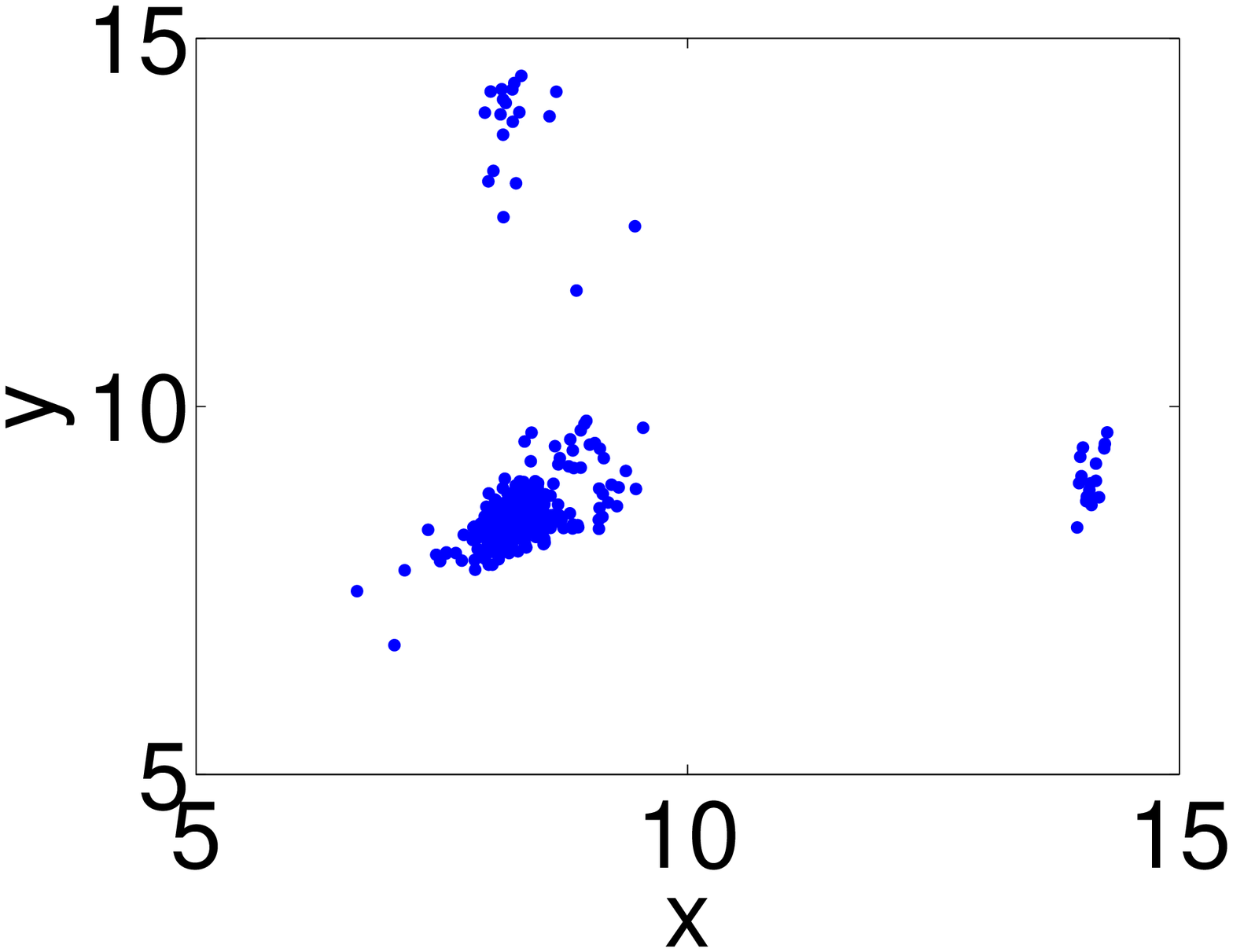}
    }
  \subfigure[lacI vs.~lacY]{
   \includegraphics[width=0.23\textwidth]{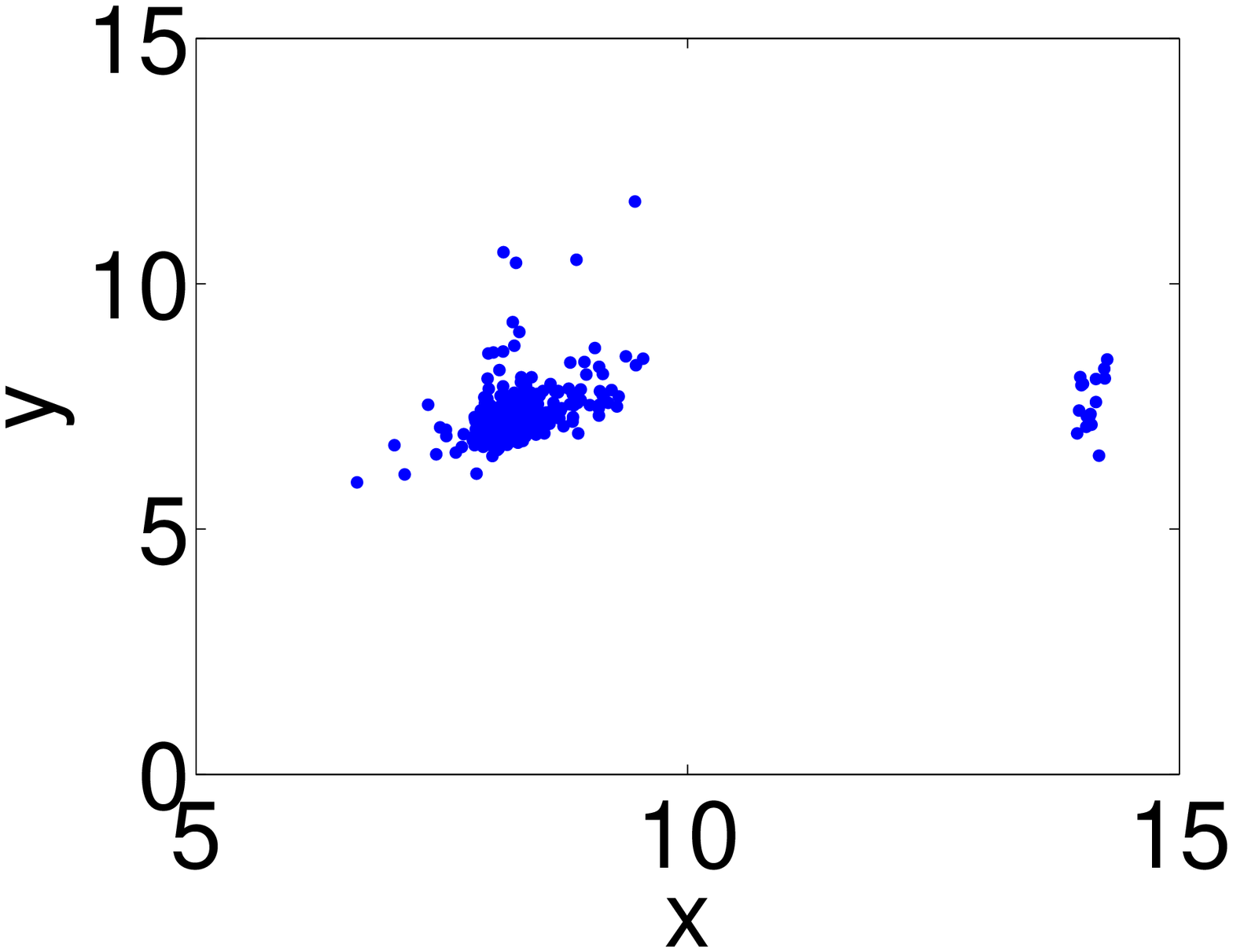}
    }
 \caption{Datasets of the E.~coli task \citep{PLOSB:Faith:2007}.}
    \label{fig:Ecoli}
\end{figure}

\subsection{Gene Function Regulations}
Finally, we apply our proposed LSIR method to the real-world biological datasets, which contain known causal
relationships about gene function regulations from transcription factors to gene expressions.

Causal prediction is biologically and medically important
because it gives us a clue for disease-causing genes or drug-target genes. Transcription factors regulate expression levels of their relating genes. In other words, when the expression level of transcription factor genes is high, genes regulated by the transcription factor become highly expressed or
suppressed.

In this experiment, we select 10 well-known gene regulation relationships of
\emph{E.~coli} \citep{PLOSB:Faith:2007}, where each data contains expression levels of the genes over 445 different environments (i.e., 445 samples, see Figure~\ref{fig:Ecoli})

The experimental results are summarized in Table~\ref{tb:Biology_causal},
showing that LSIR successfully found the correct causal direction for $7$ out of $10$ cases, while HSICR gave the correct decision only for $4$ out of $10$ cases. Moreover, the causal direction can be efficiently chosen 9 out of 10 cases just by comparing the values of $\widehat{\textnormal{SMI}}$. 


\begin{table}[t]
\centering
\caption{Results for the `E.~coli' task.
 When the $p$-values of both directions are less than $10^{-3}$,
 we concluded that the causal direction cannot be determined (indicated by `?').
 Estimated directions in the brackets are determined
 based on comparing the values
 of $\widehat{\textnormal{SMI}}$ or $\widehat{\textnormal{HSIC}}$.
}
\vspace*{2mm}
\label{tb:Biology_causal}
{
(a) LSIR\\
\begin{tabular}{|c|c||c|c|c|c|cc|c|}
\hline
\multicolumn{2}{|c||}{Dataset} & \multicolumn{2}{c|}{$p$-values} &  \multicolumn{2}{c|}{$\widehat{\textnormal{SMI}}$}  &
\multicolumn{3}{c|}{Direction}\\ \cline{1-9}
$X$ & $Y$ & $X\rightarrow Y$ & $X \leftarrow Y$ & $X\rightarrow Y$ & $X \leftarrow Y$ 
& \multicolumn{2}{|c|}{Estimated} & Truth\\
\hline\hline
lexA & uvrA &  $< 10^{-3} $ & $< 10^{-3} $  & 0.0177  & 0.0255& ? &($\rightarrow$)& $\rightarrow$\\ 
lexA & recA &  0.024        & 0.061         & 0.0070  & 0.0053& $\leftarrow$&($\leftarrow$)& $\rightarrow$\\ 
lexA & uvrB &  $< 10^{-3} $ & $< 10^{-3} $  & 0.0172  & 0.0356& ? &($\rightarrow$)& $\rightarrow$\\ 
lexA & uvrD &  0.043        & $< 10^{-3} $  & 0.0075  & 0.0227& $\rightarrow$&($\rightarrow$)& $\rightarrow$\\ 
crp  & lacA &  0.143        &  $< 10^{-3} $ & -0.0004 & 0.0399& $\rightarrow$&($\rightarrow$)& $\rightarrow$\\ 
crp  & lacY &  0.003        & $< 10^{-3} $  & 0.0118  & 0.0247& $\rightarrow$&($\rightarrow$) & $\rightarrow$\\ 
crp  & lacZ &  0.001        &  $< 10^{-3} $ & 0.0122  & 0.0307& $\rightarrow$&($\rightarrow$) & $\rightarrow$\\ 
lacI & lacA &  0.787        & $< 10^{-3} $  & -0.0076 & 0.0184& $\rightarrow$&($\rightarrow$) & $\rightarrow$\\ 
lacI & lacZ &  0.002        &  $< 10^{-3} $ & 0.0096  & 0.0141& $\rightarrow$&($\rightarrow$)& $\rightarrow$\\ 
lacI & lacY &  0.746        & $< 10^{-3} $  & -0.0082& 0.0217 &  $\rightarrow$ &($\rightarrow$)& $\rightarrow$\\ \hline
\end{tabular} \\
\vspace{0.3cm}
(b) HSICR\\
\begin{tabular}{|c|c||c|c|c|c|cc|c|}
\hline
\multicolumn{2}{|c||}{Dataset} & \multicolumn{2}{c|}{$p$-values}  &  \multicolumn{2}{c|}{$\widehat{\textnormal{HSIC}}$}  &
\multicolumn{3}{c|}{Direction}\\ \cline{1-9}
$X$ & $Y$ & $X\rightarrow Y$ & $X \leftarrow Y$ & $X\rightarrow Y$ & $X \leftarrow Y$ 
& \multicolumn{2}{|c|}{Estimated} & Truth\\
\hline\hline
lexA & uvrA &  $< 10^{-3} $ & $< 10^{-3} $  & 0.0865  & 0.1990 & ? &($\rightarrow$)& $\rightarrow$\\ 
lexA & recA &  $< 10^{-3} $ & $< 10^{-3} $  & 0.2129  & 0.1625 & ? &($\leftarrow$)& $\rightarrow$\\ 
lexA & uvrB &  0.005        & $< 10^{-3} $  & 0.0446  & 0.1335& $\rightarrow$ &($\rightarrow$)& $\rightarrow$\\ 
lexA & uvrD &  $< 10^{-3} $        & $< 10^{-3} $  & 0.0856  & 0.2427& ?&($\rightarrow$)& $\rightarrow$\\ 
crp  & lacA &  0.006        &  $< 10^{-3} $ & 0.0362 & 0.1162& $\rightarrow$&($\rightarrow$)& $\rightarrow$\\ 
crp  & lacY &  $< 10^{-3} $        & $< 10^{-3} $  & 0.0393  & 0.1303& ? &($\rightarrow$)& $\rightarrow$\\ 
crp  & lacZ &  $< 10^{-3} $ &  $< 10^{-3} $ & 0.0832  & 0.0836 & ? &($\rightarrow$)& $\rightarrow$\\ 
lacI & lacA &  0.004        & $< 10^{-3} $  & 0.0368 & 0.1076 & $\rightarrow$ &($\rightarrow$)& $\rightarrow$\\ 
lacI & lacZ &  $< 10^{-3} $        &  $< 10^{-3} $ & 0.0666  & 0.1365 & ?&($\rightarrow$)& $\rightarrow$\\ 
lacI & lacY &  0.026        & $< 10^{-3} $  & 0.0303 & 0.0927 &  $\rightarrow$ &($\rightarrow$)& $\rightarrow$\\ \hline
\end{tabular} 
}
\end{table}

\section{Conclusions}\label{sec:conclusions}
In this paper, we proposed a new method of dependence minimization regression 
called {\em least-squares independence regression} (LSIR).
LSIR adopts the \emph{squared-loss mutual information}
as an independence measure, and it is estimated by the
method of {\em least-squares mutual information} (LSMI).
Since LSMI provides an analytic-form solution, we can explicitly compute 
the gradient of the LSMI estimator with respect to regression parameters.
A notable advantage of the proposed LSIR method over
the state-of-the-art method of dependence minimization regression
\citep{ICML:Mooij+etal:2009} is that
LSIR is equipped with a natural cross-validation procedure,
allowing us to objectively optimize tuning parameters
such as the kernel width and the regularization parameter in a data-dependent fashion.
We experimentally showed that LSIR is promising
in real-world causal direction inference.

\section*{Acknowledgments}
MY was supported by the JST PRESTO program and MS was supported by SCAT, AOARD, and the FIRST program.

\bibliography{mlj_lsir}

\begin{thebibliography}{}
\bibitem[Pearson(1900)Pearson]{PhMag:Pearson:1900}
Pearson, K. (1900).
\newblock On the criterion that a given system of deviations from the probable
  in the case of a correlated system of variables is such that it can be
  reasonably supposed to have arisen from random sampling.
\newblock {\em Philosophical Magazine\/}, {\bf 50}, 157--175.
\end{thebibliography}
\bibliographystyle{natbib}
\end{document}